\documentclass{article}

\usepackage[pdftex]{graphicx, color}
\usepackage{ascmac}
\usepackage{amsthm,amsmath,amsfonts,amssymb}
\usepackage{algorithm,algorithmic}
\usepackage{bm}
\usepackage[title]{appendix}
\usepackage[colorlinks=true, linkcolor=blue, filecolor=blue, urlcolor=blue, citecolor=red]{hyperref}
\usepackage{authblk}
\usepackage{indentfirst}
\usepackage{geometry}
\usepackage{bbm}
\usepackage{multirow}

\newcommand{\argmin}{\mathop{\rm arg~min}\limits}

\allowdisplaybreaks[2]

\makeatletter
\long\def\@makecaption#1#2{%
  \normalsize
  \vskip\abovecaptionskip
  \sbox\@tempboxa{#1: #2}%
  \ifdim \wd\@tempboxa >\hsize
    #1: #2\par
  \else
    \global \@minipagefalse
    \hb@xt@\hsize{\hfil\box\@tempboxa\hfil}%
  \fi
  \vskip\belowcaptionskip}
\makeatother

\title{Deep Two-Way Matrix Reordering for Relational Data Analysis}
\author[1]{Chihiro Watanabe\thanks{chihiro\_watanabe@mist.i.u-tokyo.ac.jp}} 
\author[1,2]{Taiji Suzuki\thanks{taiji@mist.i.u-tokyo.ac.jp}}
\affil[1]{{\normalsize Graduate School of Information Science Technology, The University of Tokyo, Tokyo, Japan}}
\affil[2]{{\normalsize Center for Advanced Intelligence Project (AIP), RIKEN, Tokyo, Japan}}
\date{}

\begin{document}
\maketitle

\begin{abstract}
Matrix reordering is a task to permute the rows and columns of a given observed matrix such that the resulting reordered matrix shows meaningful or interpretable structural patterns. Most existing matrix reordering techniques share the common processes of extracting some feature representations from an observed matrix in a predefined manner, and applying matrix reordering based on it. However, in some practical cases, we do not always have prior knowledge about the structural pattern of an observed matrix. To address this problem, we propose a new matrix reordering method, called deep two-way matrix reordering (DeepTMR), using a neural network model. The trained network can automatically extract nonlinear row/column features from an observed matrix, which can then be used for matrix reordering. Moreover, the proposed DeepTMR provides the denoised mean matrix of a given observed matrix as an output of the trained network. This denoised mean matrix can be used to visualize the global structure of the reordered observed matrix. We demonstrate the effectiveness of the proposed DeepTMR by applying it to both synthetic and practical datasets. 
\end{abstract}

\section{Introduction}

Matrix reordering or seriation is a task to permute the rows and columns of a given observed matrix such that the resulting matrix shows meaningful or interpretable structural patterns \cite{Behrisch2016, Liiv2010}. Such reordering-based matrix visualization techniques provide an overview of the various practical data matrices, including gene expression data \cite{Caraux2005, Eisen1998}, document-term relationship data \cite{Berry1996}, and archaeological data \cite{Ihm2005} (e.g., the relationships between tombs and objects in Egypt \cite{Petrie1899}). In particular, we focus on the two-mode two-way matrix reordering problem, where an observed matrix or \textit{relational data matrix} $A \in \mathbb{R}^{n \times p}$ represents the relationships between two generally different objects (e.g., rows for documents and columns for words) and the permutations of rows and columns are not required to be identical, even if the row and column sizes are identical (i.e., $n=p$). 

As discussed by \cite{Behrisch2016}, most matrix reordering techniques proposed so far share the common processes of extracting ``intermediate objects'' or feature representations from an observed matrix in a predefined manner, and applying matrix reordering based on the extracted intermediate objects. For instance, under \textit{biclustering-based} methods \cite{Madeira2004, Tanay2005}, which is one of the seven categories defined by \cite{Behrisch2016}, we assume that an observed matrix consists of homogeneous submatrices or \textit{biclusters}, in which the entries are generated in an i.i.d. sense. Based on this assumption, we first estimate the locations (i.e., a set of row and column indices) of such biclusters and then reorder the rows and columns of the original matrix according to the estimated bicluster structure. In this example, the intermediate objects correspond to the bicluster assignments for the rows and columns. 

However, in some practical cases, we do not always have prior knowledge about the structural pattern of a given observed matrix, or what input features should be used as intermediate objects. In such cases, we need to apply multiple methods and compare the results to examine which method is more suitable for analyzing the given observed matrix. Therefore, the procedure of feature extraction from an observed matrix, as well as row/column reordering, should ideally be automatically fitted to a given observed matrix. 

To address this problem, we propose a new matrix reordering method, called deep two-way matrix reordering (DeepTMR), using a neural network model. The proposed DeepTMR consists of a neural network model that can be trained in an end-to-end manner and the shallower part (i.e., encoder) of the trained network can automatically extract row/column features for matrix reordering based on the given observed matrix. The expressive power of deep neural network models has been extensively studied in the literature, including the well-known universal approximation theorems \cite{Cybenko1989, Funahashi1989, Irie1988}. 
To exploit the flexibility of neural networks for feature extraction, we transform the matrix reordering problem into a parameter estimation of the neural network, which maps row and column input features to each entry value of the observed matrix. By using an autoencoder-like neural network architecture, we train the proposed DeepTMR to extract one-dimensional row/column features from a given observed matrix, such that each entry of the observed matrix can be successfully reconstructed based on the extracted features. Then, the rows and columns are reordered based on the row/column features extracted by the trained network. 

The remainder of this paper proceeds as follows. We first review the existing matrix reordering methods and describe the differences between them and the proposed DeepTMR in Section \ref{sec:related}. Then, we explain how we perform two-way matrix reordering using the proposed DeepTMR in Section \ref{sec:method}. In Section \ref{sec:experiments}, we experimentally demonstrate the effectiveness of DeepTMR by applying it to both synthetic and practical data matrices. Finally, we discuss the results and future work directions in Section \ref{sec:discussion} and conclude with Section \ref{sec:conclusion}. 


\section{Review of the studies on matrix reordering}
\label{sec:related}

According to a recent survey \cite{Behrisch2016}, matrix reordering algorithms can be roughly classified into seven categories: \textit{Robinsonian}, \textit{spectral}, \textit{dimension-reduction}, \textit{heuristic}, \textit{graph-theoretic}, \textit{biclustering}, and \textit{interactive-user-controlled}. Among these, we refer to the spectral and dimension-reduction methods, which are based on singular value decomposition (SVD) \cite{Friendly2002, Liu2003} and multidimensional scaling (MDS) \cite{Rodgers1992, Spence1974}. These methods are particularly relevant to the proposed DeepTMR in that we assume a low-dimensional latent structure for an observed matrix. In the Robinsonian and graph-theoretic methods, the general purpose is to identify the optimal row/column orders for a given loss function \cite{BarJoseph2001, Brusco2000, Diaz2002, Robinson1951, Zhao2020}. However, as to obtain the global optimal solution for such a combinatorial optimization problem becomes infeasible with increasing matrix size, we need approximated algorithms for outputting local optimal solutions. For instance, finding the optimal node reordering solution for a given arbitrary graph based on bandwidth minimization or profile minimization has been shown to be NP-hard \cite{Leung1984, Lin1994}. 

Conversely, under the spectral and dimension-reduction methods (as well as the proposed DeepTMR), instead of formulating matrix reordering as a combinatorial optimization problem, we assume that an observed matrix can be well approximated by a model with a low-dimensional latent structure, estimate the parameter of this model, and interpret the estimation result as a feature of matrix reordering. By this formulation, we can avoid directly solving a combinatorial optimization problem on row/column reordering. It must be noted that biclustering-based methods are based on such a low dimensionality assumption. However, unlike the proposed DeepTMR and the dimension-reduction-based methods, they focus on detecting biclusters (i.e., a set of submatrices with coherent patterns) for an observed matrix, where the row/column orders within a bicluster are not considered in general, as also pointed out by \cite{Friendly2002}. 

Another advantage of the spectral and dimension-reduction methods is that some methods, including the proposed DeepTMR, can be used to extract the ``denoised'' mean information of a given observed matrix $A \in \mathbb{R}^{n \times p}$. Assuming that an observed matrix is generated from a statistical model, the true purpose of matrix reordering is to reveal the row/column orders of the denoised mean matrix, not to maximize the similarities between the adjacent rows and columns in the original data matrix with noise. The spectral and dimension-reduction methods address this problem, whereas the Robinsonian and graph-theoretic methods do not. In particular, in the following examples, the SVD-based method \cite{Liu2003} (as well as the proposed DeepTMR) provides us with a denoised mean matrix of a given relational data matrix. For instance, based on the method of \cite{Liu2003}, we derive a rank-one approximation of the original matrix $A = \bm{r} \bm{c}^{\top}$, where $\bm{r} \in \mathbb{R}^n$ and $\bm{c} \in \mathbb{R}^p$. In this case, we can expect that approximated observed matrix $\bm{r} \bm{c}^{\top}$ preserves the global structure of original matrix $A$, whereas the noise in each entry is removed. Such a denoised mean matrix can be used to visualize the global structure of an observed matrix, together with the reordered data matrix. 

In the following two paragraphs, we refer to the basic matrix reordering methods based on SVD and MDS. Each of these methods is based on a specific assumption regarding the low-dimensionality of an observed matrix. The main advantages of the proposed DeepTMR for these conventional methods are as follows. 
\begin{itemize}
\item The proposed DeepTMR can extract the low-dimensional row/column features from an observed matrix more flexibly than other methods. Unlike SVD-based methods, DeepTMR can be applied without a bilinear assumption. Moreover, unlike MDS, it does not require the specification of a distance function in advance to appropriately represent the relationships (i.e., proximity) between the pairs of rows/columns. The row/column encoder of DeepTMR, which applies a nonlinear mapping from a row/column to a one-dimensional feature, is automatically obtained by training a neural network model. 
\item Unlike MDS, DeepTMR can provide us with the denoised mean matrix of a given observed matrix, as well as row/column orders. Such a denoised mean matrix can be obtained as an output of the trained neural network and can be used to visualize the global structure of the reordered observed matrix. 
\end{itemize}

\paragraph{SVD - (1) Rank-one approximation (SVD-Rank-One)} Several studies have proposed utilizing SVD for matrix reordering \cite{Friendly2002, Liu2003}. For instance, \cite{Liu2003} have proposed to model an $n \times p$ observed matrix $A$ with the following bilinear form: 
\begin{align}
&\bm{r} = (r_i)_{1 \leq i \leq n}, \ \ \ \bm{c} = (c_j)_{1 \leq j \leq p}, \ \ \ E = (E_{ij})_{1 \leq i \leq n, 1 \leq j \leq p}, \nonumber \\
&A = \bm{r} \bm{c}^{\top} + E, 
\end{align}
where $\bm{r}$ and $\bm{c}$ are the parameters corresponding to the rows and columns, respectively, and $E$ is a residual matrix. Based on the above model, we estimate parameters $\bm{r}$ and $\bm{c}$ as follows: 
\begin{align}
\label{eq:bilinear}
&\bm{\theta} \equiv \begin{bmatrix} r_1 & \cdots & r_n & c_1 & \cdots & c_p \end{bmatrix}^{\top}, \nonumber \\
&\hat{\bm{\theta}} = \argmin_{\bm{\theta} \in \mathbb{R}^{n+p}} \| A - \bm{r} \bm{c}^{\top} \|_{\mathrm{F}}^2. 
\end{align}
It can be proven that the optimal solution of (\ref{eq:bilinear}) is given by $\hat{\bm{r}} = \sqrt{\lambda_1} \bm{u}_1$ and $\hat{\bm{c}} = \sqrt{\lambda_1} \bm{v}_1$, where $\lambda_1$ is the largest singular value of matrix $A$ and $\bm{u}_1 \in \mathbb{R}^n$ and $\bm{v}_1 \in \mathbb{R}^p$ are the corresponding row and column singular vectors, respectively. Therefore, the order of the estimated row and column parameters $\hat{\bm{r}}$ and $\hat{\bm{c}}$ can be respectively used for matrix reordering. 

\paragraph{SVD - (2) Angle between the top two singular vectors (SVD-Angle)} Friendly \cite{Friendly2002} has pointed out that the structure of a given data matrix cannot always be represented sufficiently by a single principal component and has proposed to define the row/column orders using the angle between the top two singular vectors. Under this method, an observed matrix $A$ is first mean-centered and scaled as follows: 
\begin{align}
&\tilde{A}^{(0)} = (\tilde{A}^{(0)}_{ij})_{1 \leq i \leq n, 1 \leq j \leq p}, \ \ \ \tilde{A}^{(0)}_{ij} = A_{ij} - \frac{1}{p} \sum_{j = 1}^p A_{ij}, \nonumber \\
&\tilde{A} = (\tilde{A}_{ij})_{1 \leq i \leq n, 1 \leq j \leq p}, \ \ \ 
\tilde{A}_{ij} = \frac{\tilde{A}^{(0)}_{ij}}{\sqrt{\frac{1}{p} \sum_{j = 1}^p \left( \tilde{A}^{(0)}_{ij} \right)^2}}. 
\end{align}
Let $\bm{u}_1$ and $\bm{u}_2$ be the row singular vectors of scaled observed matrix $\tilde{A}$ which correspond to the largest and second largest singular values. Angle $\alpha_i$ between the top two row singular vectors is given by: 
\begin{align}
\alpha_i = \tan^{-1} (u_{i2} / u_{i1}) + \pi I[u_{i1} \leq 0], 
\end{align}
where $I[\cdot]$ is an indicator function. The row order is determined by splitting angles $\{ \alpha_i \}$ at the largest gap between two adjacent angles. The column order can then be defined in the same way as the row one, by replacing observed matrix $\tilde{A}$ with transposed matrix $\tilde{A}^{\top}$. 

\paragraph{MDS} MDS is also a dimension reduction method that can be used for matrix reordering \cite{Rodgers1992, Spence1974}. Under MDS, we use a proximity matrix, each entry representing the distance between a pair of rows or columns. For instance, we can define a proximity matrix $D$ for rows based on a given observed matrix $A \in \mathbb{R}^{n \times p}$ as follows: 
\begin{align}
D = (D_{ii'})_{1 \leq i, i' \leq n}, \ \ \ 
D_{ii'} = \left( \sum_{j = 1}^n (A_{ij} - A_{i'j})^2 \right)^{\frac{1}{2}}, \ \ \ 
i, i' = 1, \dots, n. 
\end{align}
The purpose of MDS is to obtain a $k$-dimensional representation, $\tilde{A} \in \mathbb{R}^{n \times k}$, of the original observed matrix, $A$, based on matrix $D$, where $k \leq n, p$. First, we define the following matrices: 
\begin{align}
&\tilde{D} = (\tilde{D}_{ii'})_{1 \leq i, i' \leq n}, \ \ \ \tilde{D}_{ii'} = D_{ii'}^2, \ \ \ i, i' = 1, \dots, n, \nonumber \\
&Q = (Q_{ii'})_{1 \leq i, i' \leq n}, \ \ \ Q_{ii'} = 1, \ \ \ i, i' = 1, \dots, n, \nonumber \\
&B = -\frac{1}{2} \left( I - n^{-1} Q \right) \tilde{D} \left( I - n^{-1} Q \right). 
\end{align}
It can be easily shown that $B$ is a semi-positive definite matrix. Let $\lambda_i$ and $\bm{v}_i$ be the $i$th largest eigenvalue of matrix $B$ and the corresponding eigenvector, respectively. The $k$-dimensional representation $\tilde{A}$ of matrix $A$ is given by: 
\begin{align}
\tilde{A} = \begin{bmatrix}
\bm{v}_1 & \cdots & \bm{v}_k
\end{bmatrix} \mathrm{diag} (\sqrt{\lambda_1}, \dots, \sqrt{\lambda_k}). 
\end{align}
It has been proven that solution $\tilde{A}$ minimizes the \textit{Strain}, which is given by $\mathcal{L} (\tilde{A}) = \| \tilde{A} \tilde{A}^{\top} - B \|_{\mathrm{F}}^2$ \cite{Borg1997}. 
By setting $k=1$, we can obtain the one-dimensional row feature of observed matrix $A$, which can be used to determine the row order. The column order can be defined in the same way as the row one, by replacing observed matrix $A$ with transposed matrix $A^{\top}$. 


\section{Deep two-way matrix reordering}
\label{sec:method}

\begin{figure}[p]
  \centering
  \includegraphics[width=0.27\hsize]{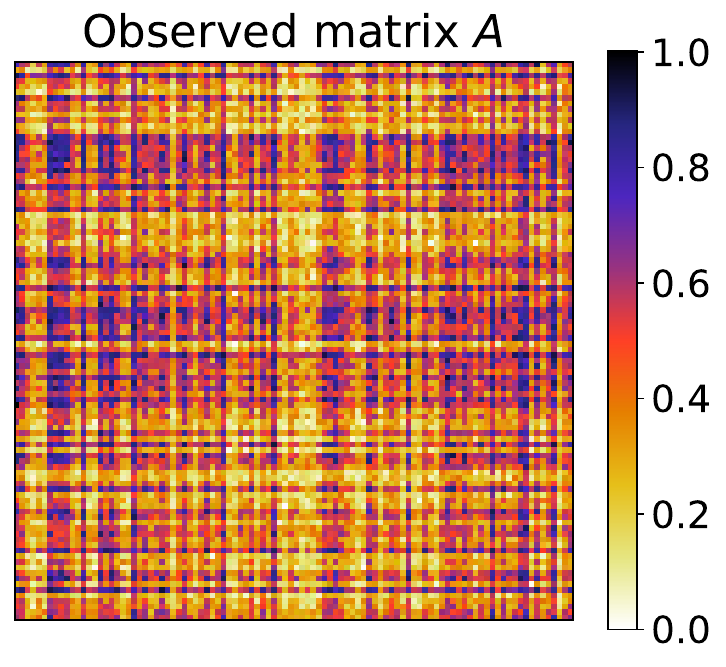}
  \includegraphics[width=0.12\hsize]{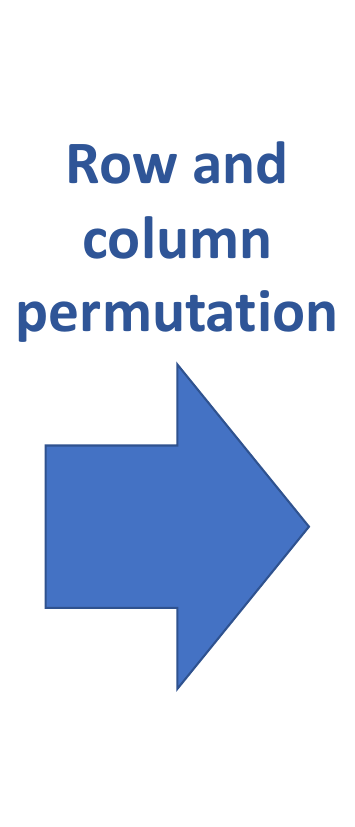}
  \includegraphics[width=0.27\hsize]{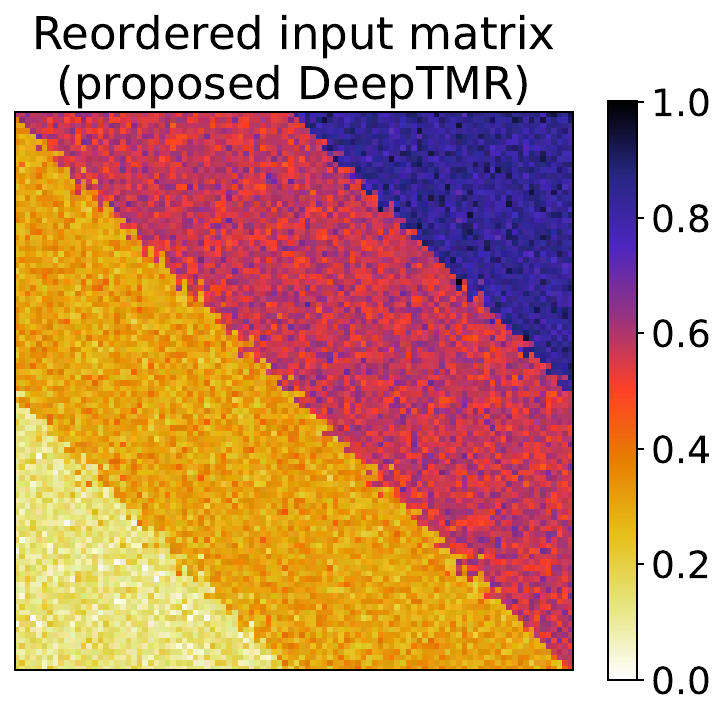}
  \includegraphics[width=0.27\hsize]{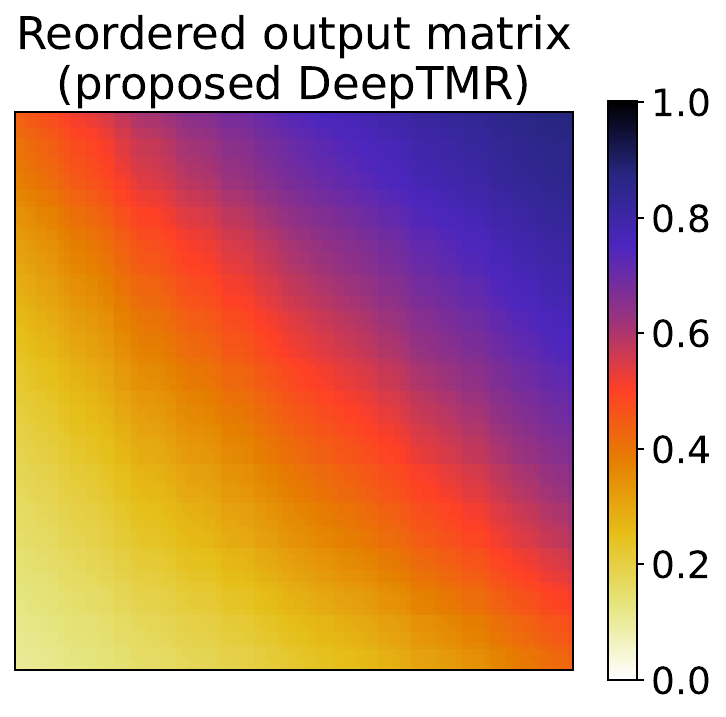}
  \caption{Matrix reordering problem. Given an observed matrix $A$ (left), the proposed DeepTMR reorders the rows and columns of matrix $A$ such that the reordered input matrix (center) shows a meaningful or interpretable structure. Th proposed DeepTMR provides us with the denoised mean matrix of the reordered matrix (right) as the output of a trained network, as well as row/column ordering.}\vspace{3mm}
  \label{fig:matrix_reordering}
  \includegraphics[width=0.75\hsize]{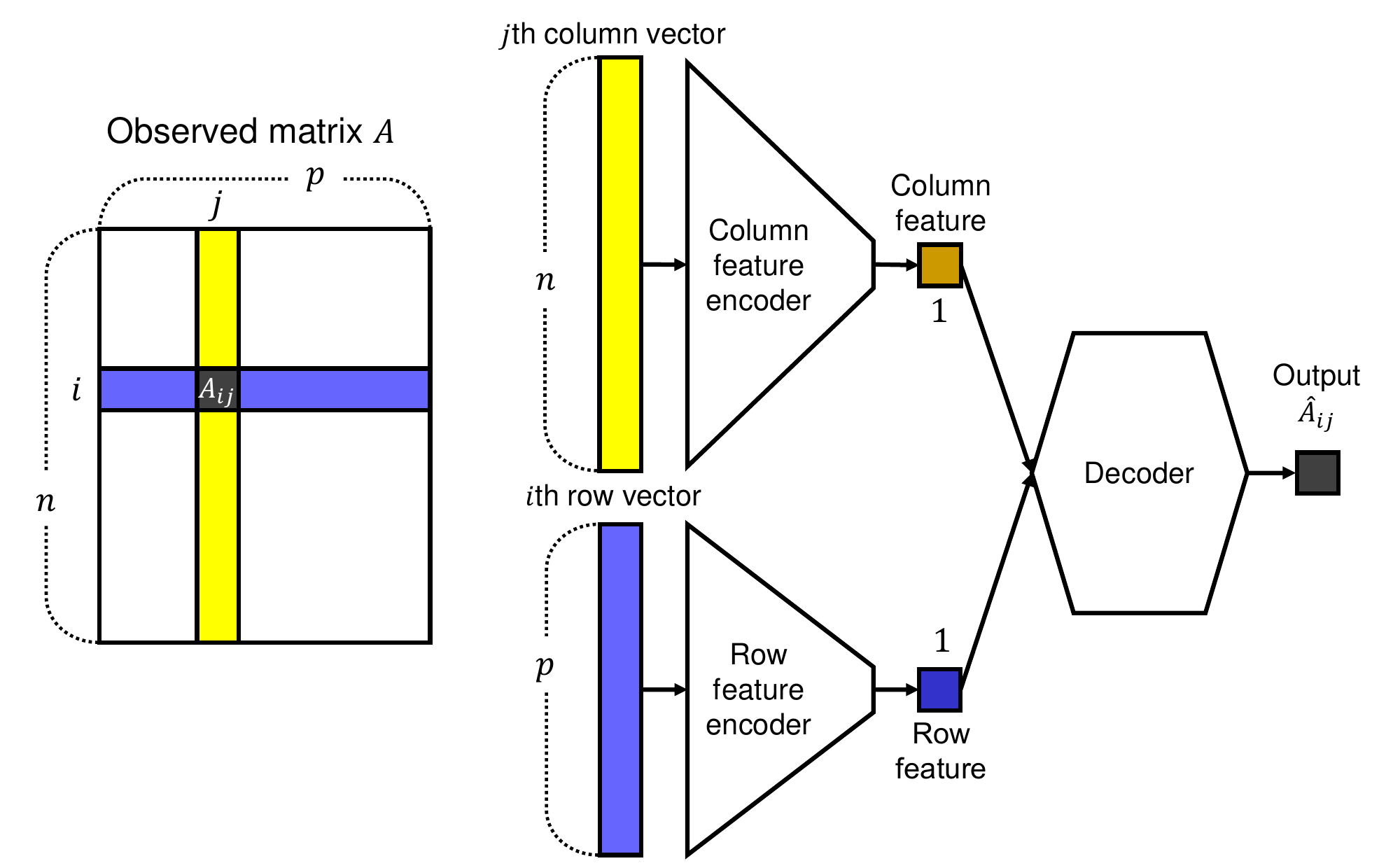}
  \caption{Model architecture of DeepTMR. Given an observed matrix $A$, DeepTMR is trained to reconstruct each entry $A_{ij}$ from one-dimensional row and column features, which are extracted from the $i$th row and the $j$th column of matrix $A$. After training the network, we reorder the rows and columns of matrix $A$ based on the row and column features extracted in the middle layer.}
  \label{fig:architecture}
\end{figure}

Given an $n \times p$ observed matrix $A \in \mathbb{R}^{n \times p}$, our purpose is to reorder the row and column indices of matrix $A$ such that the resulting matrix, $\underline{A}$, exhibits some structure (e.g., block structure), as shown in Figure \ref{fig:matrix_reordering}. 

Figure \ref{fig:architecture} shows the entire network architecture of the proposed DeepTMR. To extract the row and column features of given matrix $A$, we propose a new neural network model, DeepTMR, which has an autoencoder-like architecture. Under DeepTMR, the $(i, j)$th entry $A_{ij}$ of the observed matrix $A$ is estimated based on its row and column data vectors, $\bm{r}^{(i)}$ and $\bm{c}^{(j)}$, respectively, which are given by: 
\begin{align}
\bm{r}^{(i)} = (r^{(i)}_{j'})_{1 \leq j' \leq p}, \ \ \ r^{(i)}_{j'} = A_{ij'}, 
\nonumber \\
\bm{c}^{(j)} = (c^{(j)}_{i'})_{1 \leq i' \leq n}, \ \ \ c^{(j)}_{i'} = A_{i'j}. 
\end{align}

Then, from these input data vectors, the features of the $i$th row and the $j$th column, $g_i$ and $h_j$, respectively, are extracted by row and column encoder networks: 
\begin{align}
g_i = \mathrm{ROWENC} \left( \bm{r}^{(i)} \right), \\
h_j = \mathrm{COLUMNENC} \left( \bm{c}^{(j)} \right). 
\end{align}
Here, $\mathrm{ROWENC}: \mathbb{R}^p \mapsto \mathbb{R}$ and $\mathrm{COLUMNENC} (\cdot): \mathbb{R}^n \mapsto \mathbb{R}$ can be implemented as arbitrary neural network architectures, provided that they have a fixed number of units $m$ and $\tilde{m}$ in the input and output layers, respectively, that is, $(m, \tilde{m}) = (p, 1)$ for $\mathrm{ROWENC} (\cdot)$ and $(m, \tilde{m}) = (n, 1)$ for $\mathrm{COLUMNENC} (\cdot)$. 

From these row and column features, the $(i, j)$th entry, $A_{ij}$, is estimated using a decoder network: 
\begin{align}
\hat{A}_{ij} = \mathrm{DEC} \left( g_i, h_j \right), 
\end{align}
where $\mathrm{DEC}: \mathbb{R}^2 \mapsto \mathbb{R}$ can be implemented as an arbitrary neural network architecture with two input layer units and one output layer unit. 

By using mini-batch learning, the entire network is trained such that the following mean squared error with the $L_2$ regularization term is minimized: 
\begin{align}
\mathcal{L} = \frac{1}{|\mathcal{I}_t|} \sum_{(i, j) \in \mathcal{I}_t} (A_{ij} - \hat{A}_{ij})^2 + \lambda \| \bm{w} \|_2^2, 
\end{align}
where $\mathcal{I}_t$ is a set of row and column indices $(i, j)$ in a mini-batch of the $t$th iteration, $\lambda$ is a hyperparameter, and $\bm{w}$ is a vector of parameters for the entire network\footnote{In the experiments in Section \ref{sec:experiments}, we define $\bm{w}$ as a vector of all weights and biases in the linear layers of the encoder and decoder networks.}. 

Finally, we define matrix $\underline{A}$ with the reordered rows and columns. By using the trained row and column encoder networks, we define the following two feature vectors: 
\begin{align}
\bm{g} &= \begin{bmatrix} \mathrm{ROWENC} \left( \bm{r}^{(1)} \right) & \dots & \mathrm{ROWENC} \left( \bm{r}^{(n)} \right) \end{bmatrix}^{\top}, \nonumber \\
\bm{h} &= \begin{bmatrix} \mathrm{COLUMNENC} \left( \bm{c}^{(1)} \right) & \dots & \mathrm{COLUMNENC} \left( \bm{c}^{(p)} \right) \end{bmatrix}^{\top}.
\end{align}
Because the network has been trained to recover each entry value from only the corresponding row and column data vectors, vectors $\bm{g}$ and $\bm{h}$ can be expected to reflect the row and column features of the original matrix, $A$. Based on this conjecture, we define $\pi^{\mathrm{row}}$ as a permutation of $\{ 1, 2, \dots, n \}$ that represents the ascending order of the entries of $\bm{g}$ (i.e., $g_{\pi^{\mathrm{row}} (1)} \leq g_{\pi^{\mathrm{row}} (2)} \leq \dots \leq g_{\pi^{\mathrm{row}} (n)}$ holds). Similarly, we define $\pi^{\mathrm{column}}$ as a permutation of $\{ 1, 2, \dots, p \}$ representing the ascending order of the entries of $\bm{h}$ (i.e., $h_{\pi^{\mathrm{column}} (1)} \leq h_{\pi^{\mathrm{column}} (2)} \leq \dots \leq h_{\pi^{\mathrm{column}} (p)}$ holds). Using these row and column permutations, we respectively obtain the reordered row and column features, $\underline{\bm{g}}$ and $\underline{\bm{h}}$, and the reordered observed and estimated matrices, $\underline{A}$ and $\underline{\hat{A}}$, as follows: 
\begin{align}
&\underline{\bm{g}} = (\underline{g}_i)_{1 \leq i \leq n}, \ \ \ \underline{g}_i = g_{\pi^{\mathrm{row}} (i)}, \nonumber \\
&\underline{\bm{h}} = (\underline{h}_j)_{1 \leq j \leq p}, \ \ \ \underline{h}_j = h_{\pi^{\mathrm{column}} (j)}, \nonumber \\
&\underline{A} = (\underline{A}_{ij})_{1 \leq i \leq n, 1 \leq j \leq p}, \ \ \ \underline{A}_{ij} = A_{\pi^{\mathrm{row}} (i) \pi^{\mathrm{column}} (j)}, \nonumber \\
&\underline{\hat{A}} = (\underline{\hat{A}}_{ij})_{1 \leq i \leq n, 1 \leq j \leq p}, \ \ \ \underline{\hat{A}}_{ij} = \hat{A}_{\pi^{\mathrm{row}} (i) \pi^{\mathrm{column}} (j)}. 
\end{align}


\section{Experiments}
\label{sec:experiments}

To verify the effectiveness of DeepTMR, we applied it to both synthetic and practical relational data matrices and plotted their latent row-column structures. For all the experiments: 
\begin{itemize}
\item We initialized the weights and biases of the linear layers using the method described by \cite{Glorot2010}. In other words, each weight value that connects the $l$th and $(l+1)$th layers is initialized based on a uniform distribution on interval $\left[ -1/\sqrt{m^{(l)}}, 1/\sqrt{m^{(l)}} \right]$, where $m^{(l)}$ is the number of units in the $l$th layer. As for the biases, we set their initial values to zero. 
\item We used the Adam optimizer \cite{Kingma2015} with $\beta_1 = 0.9$, $\beta_2 = 0.999$, and $\epsilon = 1.0 \times 10^{-8}$ for training the DeepTMR network\footnote{As for learning rates $\eta$, we used different settings for each experiment, as shown in Table \ref{tab:hyperparameter}}. 
\end{itemize}

\subsection{Preliminary experiment using synthetic datasets}
\label{sec:preliminary_exp}

First, we generated three types of synthetic datasets with latent row and column structures, applied the DeepTMR, and checked whether we could successfully recover the latent structure of the given observed matrices. For all three models, we set the matrix size to $(n, p) = (100, 100)$. 

\paragraph{Latent block model} First, we generated a matrix based on a latent block model (LBM) \cite{Arabie1978, Govaert2003, Hartigan1972}. Under an LBM, we assume that each row and column of a given matrix, $\bar{A}^{(0)} \in \mathbb{R}^{n \times p}$, belong to one of the $K$ row and $H$ column clusters, respectively. Let $c_i$ and $d_j$ be the row cluster index of the $i$th row and the column cluster index of the $j$th column of matrix $\bar{A}^{(0)}$. In this experiment, we set the number of row and column clusters at $(K, H) = (3, 3)$, and define the row and column cluster assignments as follows: 
\begin{align}
&n^{(0)} = \mathrm{ceil} \left( \frac{n}{K} \right), \ \ \ 
p^{(0)} = \mathrm{ceil} \left( \frac{p}{H} \right), \nonumber \\
&c_1 = \dots = c_{n^{(0)}} = 1, \ \ \ 
c_{n^{(0)}+1} = \dots = c_{2n^{(0)}} = 2, \ \ \ 
\dots, \ \ \ 
c_{(K-1) n^{(0)}+1} = \dots = c_n = K, \nonumber \\
&d_1 = \dots = d_{p^{(0)}} = 1, \ \ \ 
d_{p^{(0)}+1} = \dots = d_{2p^{(0)}} = 2, \ \ \ 
\dots, \ \ \ 
d_{(H-1) p^{(0)}+1} = \dots = d_p = H, 
\end{align}
where $\mathrm{ceil} (\cdot)$ is the ceiling function. Based on the above definitions, under an LBM, we assume that each entry of matrix $\bar{A}^{(0)}$ is independently generated from a block-wise identical distribution. Specifically, we generate each $(i, j)$th entry $\bar{A}^{(0)}_{ij}$ based on a Gaussian distribution with mean $B_{c_i d_j}$ and standard deviation $\sigma$ given by: 
\begin{align}
B = \begin{bmatrix}
0.9 & 0.4 & 0.8\\
0.1 & 0.6 & 0.2\\
0.5 & 0.3 & 0.7
\end{bmatrix}, \ \ \ 
\sigma = 0.05. 
\end{align}

\paragraph{Striped pattern model} To show that the DeepTMR can reveal a latent row-column structure that is not necessarily represented as a set of rectangular blocks, we also used the striped pattern model (SPM). An SPM is similar to an LBM, in that we assume that each entry of a given matrix $\bar{A}^{(0)} \in \mathbb{R}^{n \times p}$ belongs to one of the $K$ clusters and it is independently generated from a cluster-wise identical distribution. However, unlike an LBM, we assume that the cluster assignments show a striped pattern rather than a regular grid one. Specifically, let $c_{ij}$ be the cluster index of the $(i, j)$th entry $\bar{A}^{(0)}_{ij}$ of matrix $\bar{A}^{(0)}$. Under an SPM, the cluster assignment is given by: 
\begin{align}
&n^{(0)} = \mathrm{ceil} \left( \frac{n+p}{K} \right), \nonumber \\
&c_{ij} = \mathrm{floor} \left( \frac{i + j - 2}{n^{(0)}} \right) + 1, \ \ \ i = 1, \dots, n, \ \ \ j = 1, \dots, p. 
\end{align}
Based on the above cluster assignments with a striped pattern, we generated each $(i, j)$th entry $\bar{A}^{(0)}_{ij}$ of matrix $\bar{A}^{(0)}$ based on a Gaussian distribution with mean $b_{c_{ij}}$ and standard deviation $\sigma$ given by: 
\begin{align}
\bm{b} = \begin{bmatrix}
0.9 & 0.6 & 0.3 & 0.1
\end{bmatrix}, \ \ \ 
\sigma = 0.05. 
\end{align}

\paragraph{Gradation block model} Under the above LBM and SPM, we assume that each entry is generated from a cluster-wise identical distribution. In the gradation block model (GBM), we consider a different case, where a matrix $\bar{A}^{(0)} \in \mathbb{R}^{n \times p}$ contains a block or submatrix with a gradation (i.e., continuous) pattern. Specifically, let $I$ and $J$ be the sets of row and column indices that belong to the block with the gradation pattern. We assume that the $(i, j)$th entry $\bar{A}^{(0)}_{ij}$ of matrix $\bar{A}^{(0)}$ is generated from a Gaussian distribution with mean $B_{ij}$ and standard deviation $\sigma$, being given by: 
\begin{align}
&n^{(0)} = \mathrm{ceil} \left( \frac{n}{2} \right), \ \ \ 
p^{(0)} = \mathrm{ceil} \left( \frac{p}{2} \right), \nonumber \\
&B_{ij} = \begin{cases}
0.1 & \mathrm{if}\ \left( i > n^{(0)} \right) \cup \left( j > p^{(0)} \right), \\
\frac{0.8 (j - 1)}{p^{(0)} - 1} + 0.1 &\mathrm{if}\ \left( i \leq n^{(0)} \right) \cap \left( j \leq p^{(0)} \right), \end{cases} \ \ \ 
i = 1, \dots, n, \ \ \ j = 1, \dots, p, \nonumber \\
&\sigma = 0.05. 
\end{align}

For all the above three models, once we generated matrix $\bar{A}^{(0)}$, we could define matrix $\bar{A}$ as follows: 
\begin{align}
\label{eq:A_normalization}
\bar{A} = (\bar{A}_{ij})_{1 \leq i \leq n, 1 \leq j \leq p}, \ \ \ \bar{A}_{ij} = \frac{\bar{A}^{(0)}_{ij} - \min_{(i, j) = (1, 1), \dots, (n, p)} \bar{A}^{(0)}_{ij}}{\max_{(i, j) = (1, 1), \dots, (n, p)} \bar{A}^{(0)}_{ij} - \min_{(i, j) = (1, 1), \dots, (n, p)} \bar{A}^{(0)}_{ij}}. 
\end{align}
By definition, the maximum and minimum entries of matrix $\bar{A}$ are one and zero, respectively. 
Then, we applied random permutation to the rows and columns of matrix $\bar{A}$ to obtain observed matrix $A$. Finally, we applied the DeepTMR to observed matrix $A$ and checked whether it could recover the latent row-column structure of matrix $A$. The hyperparameter settings for training the DeepTMR are listed in Table \ref{tab:hyperparameter}. 

\begin{table}[t]
\centering
\caption{Experimental settings of learning rate $\eta$, number of epochs $T$ (the total number of iterations is given by $\mathrm{ceil} [Tnp/|\mathcal{I}|]$), regularization hyperparameter $\lambda$, number of sets of row and column indices in a mini-batch $|\mathcal{I}|$, and number of units in $\mathrm{ROWENC}$, $\mathrm{COLUMNENC}$, and $\mathrm{DEC}$ networks, $\bm{m}^{\mathrm{ROWENC}}$, $\bm{m}^{\mathrm{COLUMNENC}}$, and $\bm{m}^{\mathrm{DEC}}$, respectively (from input to output).}\vspace{3mm}
\scalebox{0.8}{
\begin{tabular}{|c||c|c|c|c|c|c|c|} \hline
& $\eta$ & $T$ & $\lambda$ & $|\mathcal{I}|$ & $\bm{m}^{\mathrm{ROWENC}}$ & $\bm{m}^{\mathrm{COLUMNENC}}$ & $\bm{m}^{\mathrm{DEC}}$ \\ \hline \hline
Sec. \ref{sec:preliminary_exp}, LBM & \multirow{6}{*}{$1.0 \times 10^{-2}$} & \multirow{3}{*}{$1 \times 10^2$} & \multirow{6}{*}{$1.0 \times 10^{-10}$} & \multirow{4}{*}{$2 \times 10^2$} & \multirow{6}{*}{$\begin{bmatrix} p, 10, 1 \end{bmatrix}$} & \multirow{6}{*}{$\begin{bmatrix} n, 10, 1 \end{bmatrix}$} & \multirow{6}{*}{$\begin{bmatrix} 2, 10, 1 \end{bmatrix}$} \\ \cline{1-1} 
Sec. \ref{sec:preliminary_exp}, SPM & & & & & & & \\ \cline{1-1}
Sec. \ref{sec:preliminary_exp}, GBM & & & & & & & \\ \cline{1-1} \cline{3-3} 
Sec. \ref{sec:comparison}, DGM & & \multirow{2}{*}{$2 \times 10^2$} & & & & & \\ \cline{1-1} \cline{5-5} 
Sec. \ref{sec:divorce_data} & & & & \multirow{2}{*}{$5 \times 10^2$} & & & \\ \cline{1-1} \cline{3-3} 
Sec. \ref{sec:traffic_data} & & $1 \times 10^2$ & & & & & \\ \hline
\end{tabular}}
\label{tab:hyperparameter}
\end{table}

\begin{figure}[p]
  \centering
  \includegraphics[width=0.24\hsize]{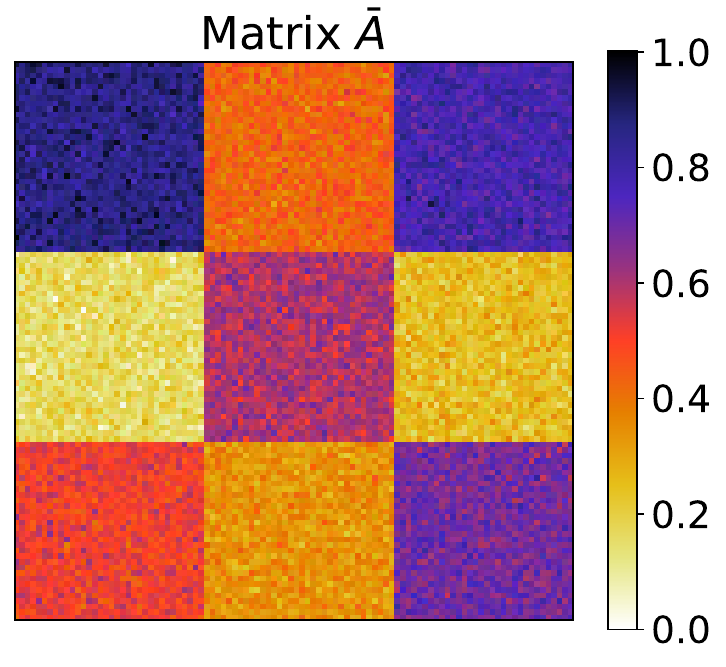}
  \includegraphics[width=0.24\hsize]{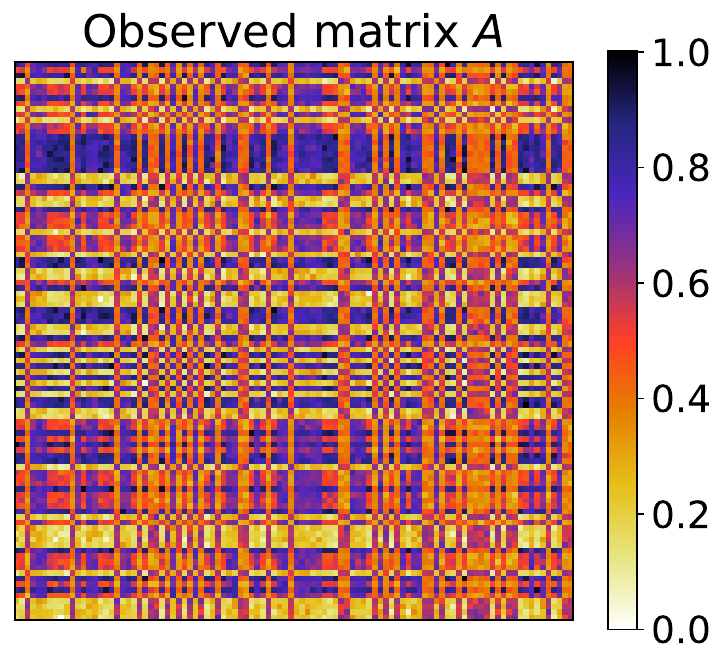}
  \includegraphics[width=0.24\hsize]{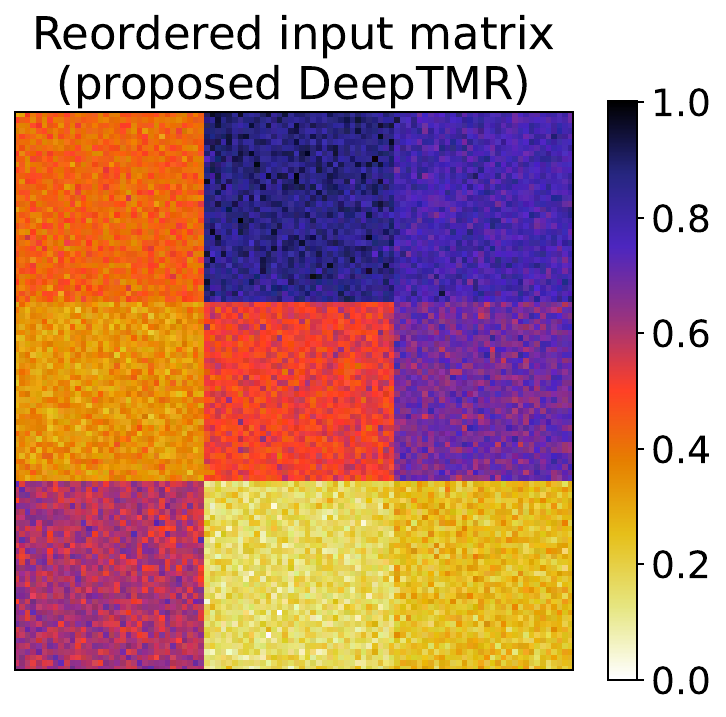}
  \includegraphics[width=0.24\hsize]{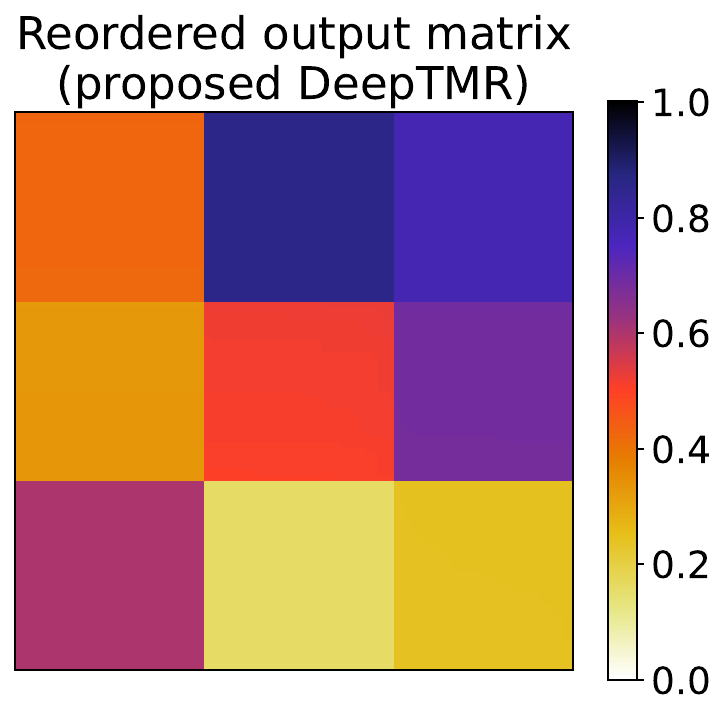}\\
  \includegraphics[width=0.45\hsize]{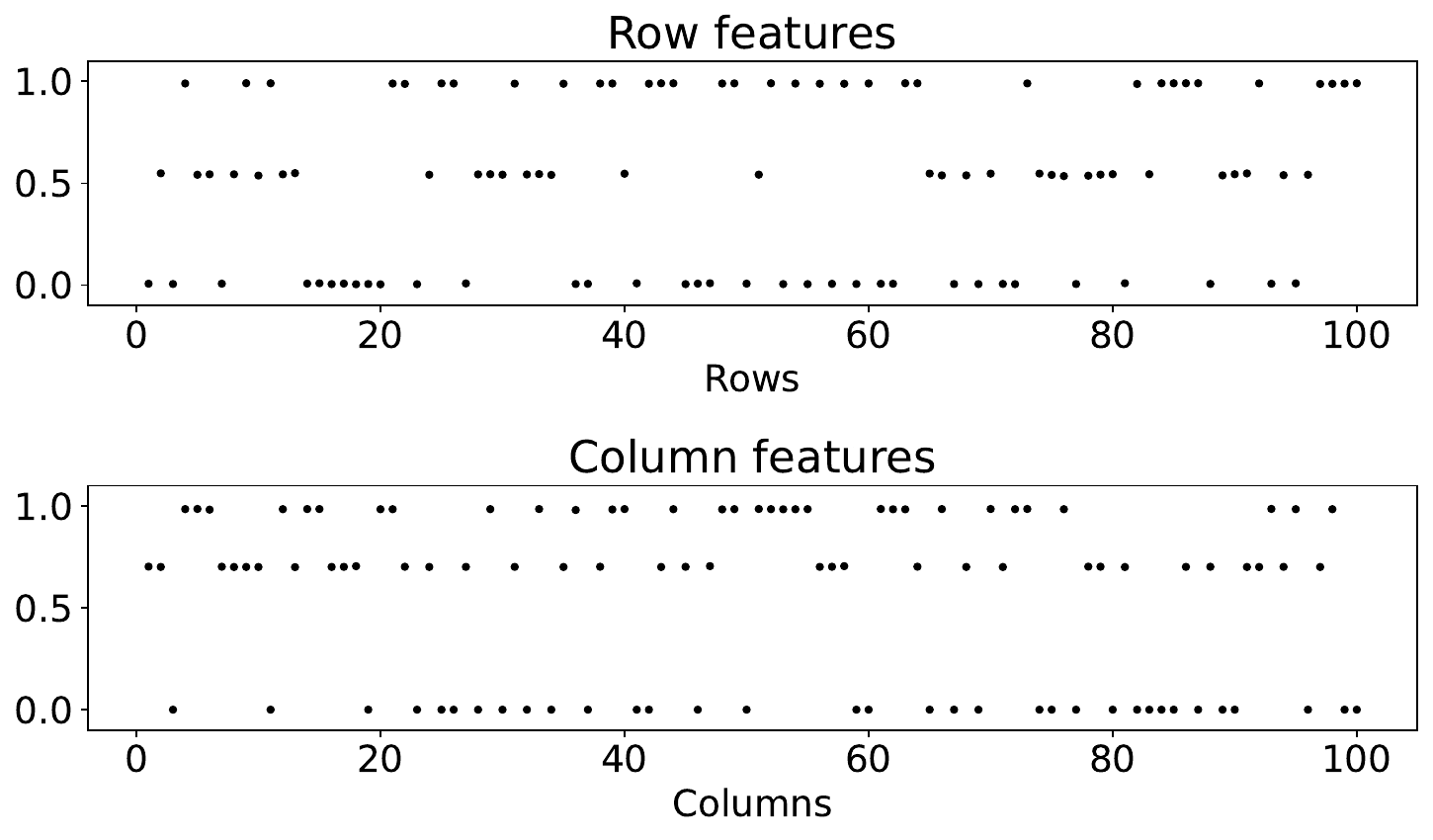}\hspace{10mm}
  \includegraphics[width=0.45\hsize]{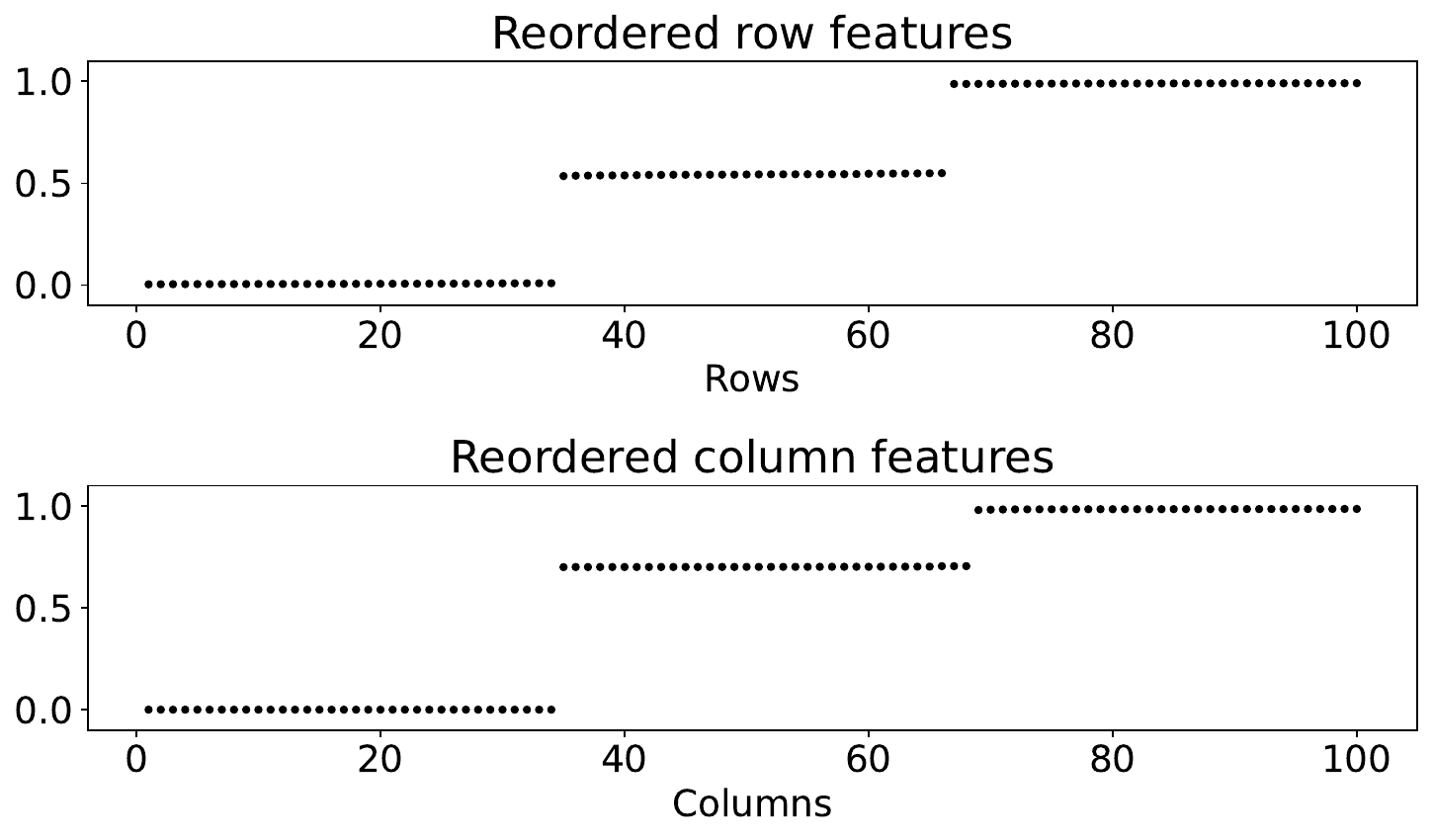}\vspace{-1mm}
  \caption{Results of the \textbf{LBM}. Top figures: original matrix $\bar{A}$, observed matrix $A$ obtained by applying random row-column permutation to $\bar{A}$, reordered input matrix $\underline{A}$, and reordered output matrix $\underline{\hat{A}}$ (left to right). Bottom figures: Encoded row and column features $\bm{g}$ and $\bm{h}$ and reordered row and column features $\underline{\bm{g}}$ and $\underline{\bm{h}}$ (left to right).}\vspace{3mm}
  \label{fig:results_m1}
  \includegraphics[width=0.24\hsize]{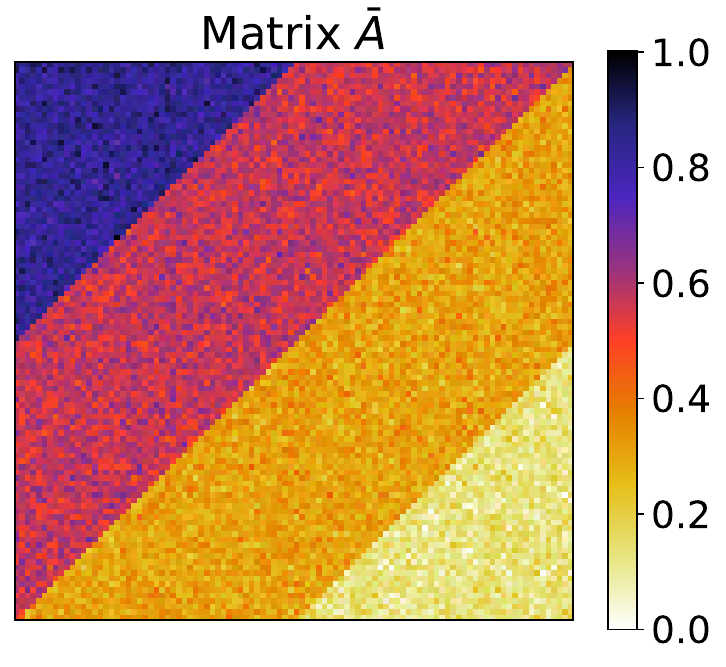}
  \includegraphics[width=0.24\hsize]{A_synthetic2_input_permutated-eps-converted-to.pdf}
  \includegraphics[width=0.24\hsize]{A_synthetic2_input_sort-eps-converted-to.pdf}
  \includegraphics[width=0.24\hsize]{A_synthetic2_out-eps-converted-to.pdf}\\
  \includegraphics[width=0.45\hsize]{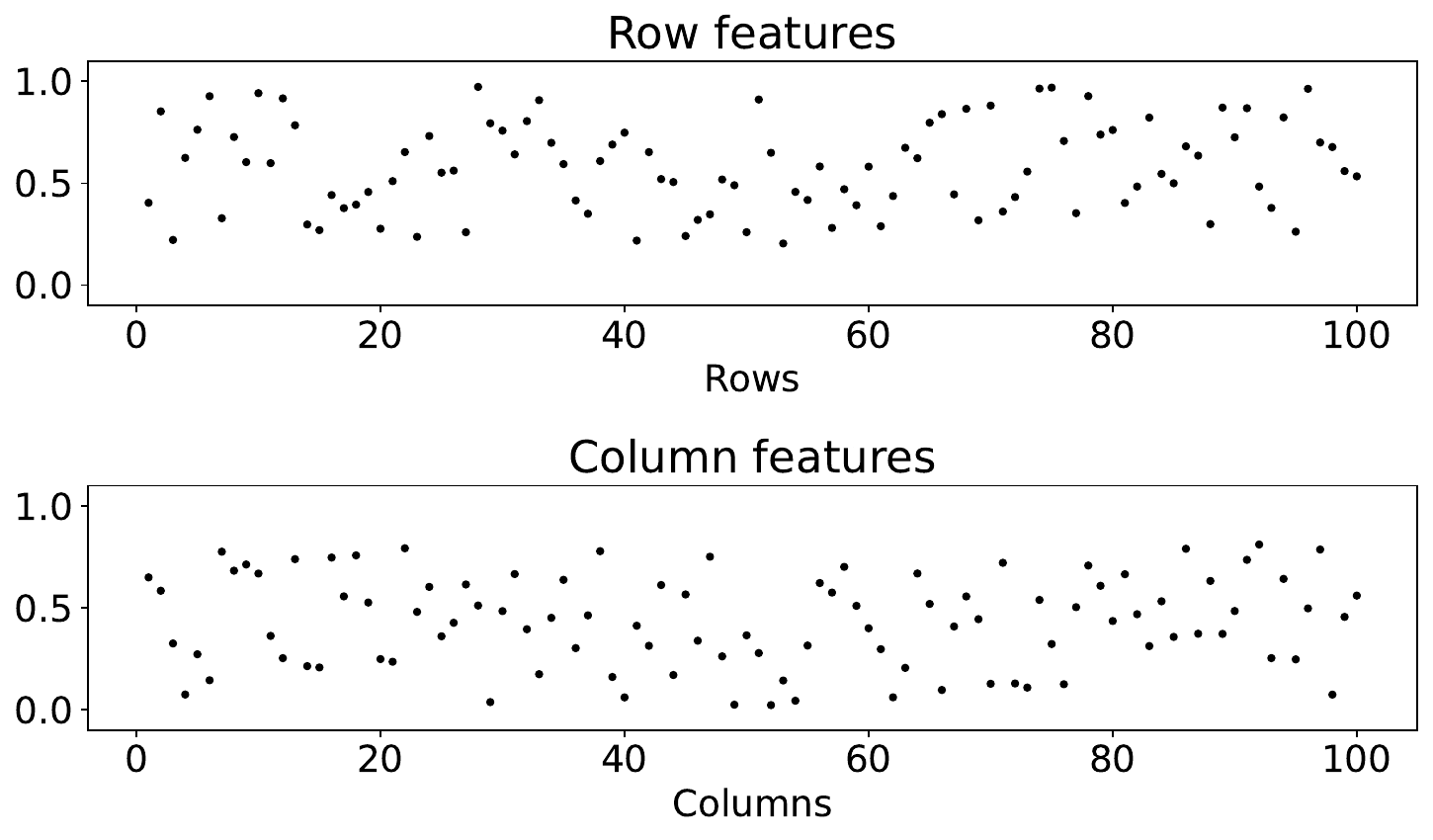}\hspace{10mm}
  \includegraphics[width=0.45\hsize]{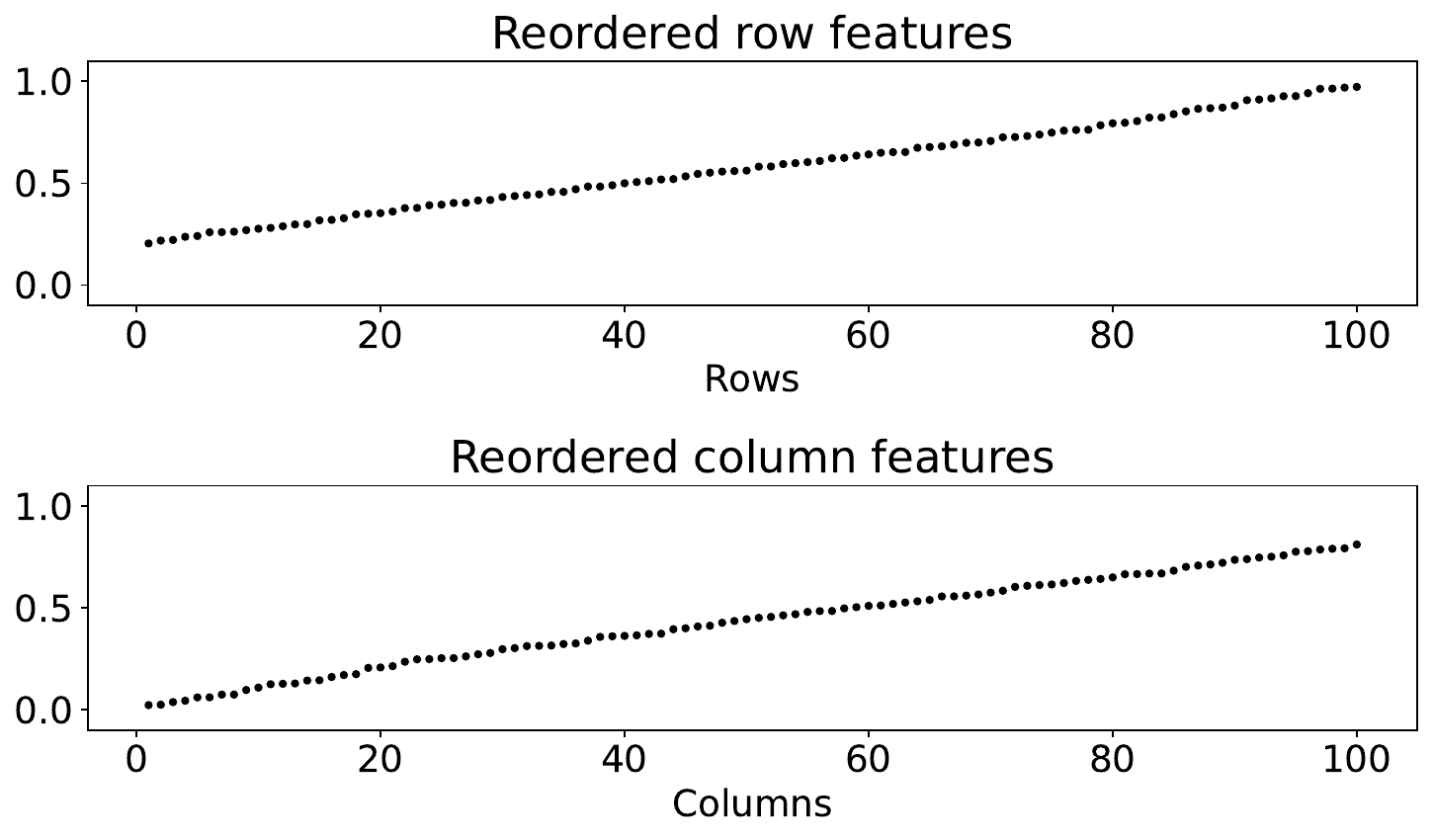}\vspace{-1mm}
  \caption{Results of the \textbf{striped pattern model} for matrices $\bar{A}$, $A$, $\underline{A}$, $\underline{\hat{A}}$, and vectors $\bm{g}$, $\bm{h}$, $\underline{\bm{g}}$, $\underline{\bm{h}}$.}\vspace{3mm}
  \label{fig:results_m2}
\end{figure}

\begin{figure}[t]
  \centering
  \includegraphics[width=0.24\hsize]{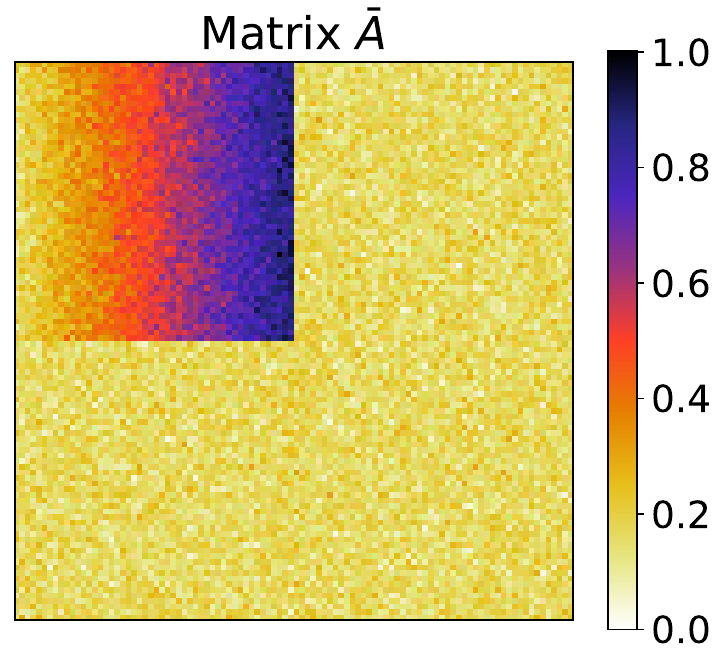}
  \includegraphics[width=0.24\hsize]{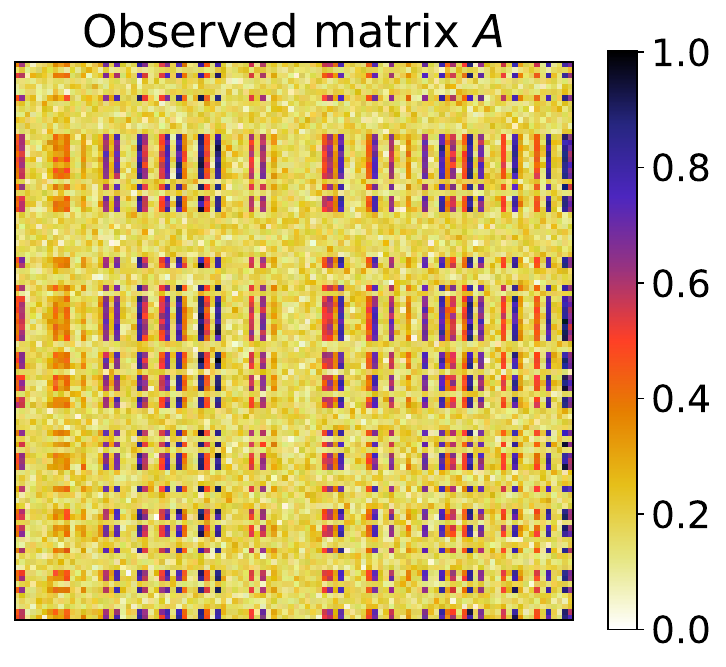}
  \includegraphics[width=0.24\hsize]{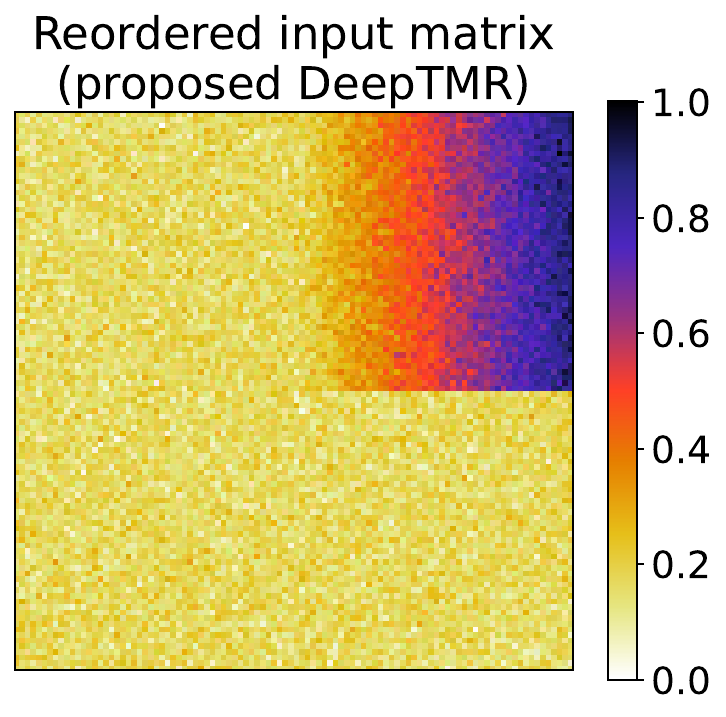}
  \includegraphics[width=0.24\hsize]{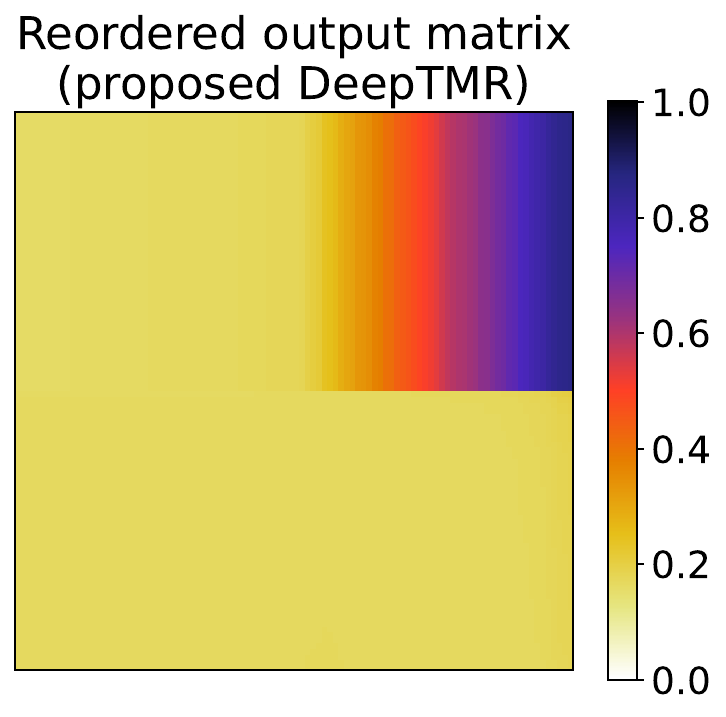}\\
  \includegraphics[width=0.45\hsize]{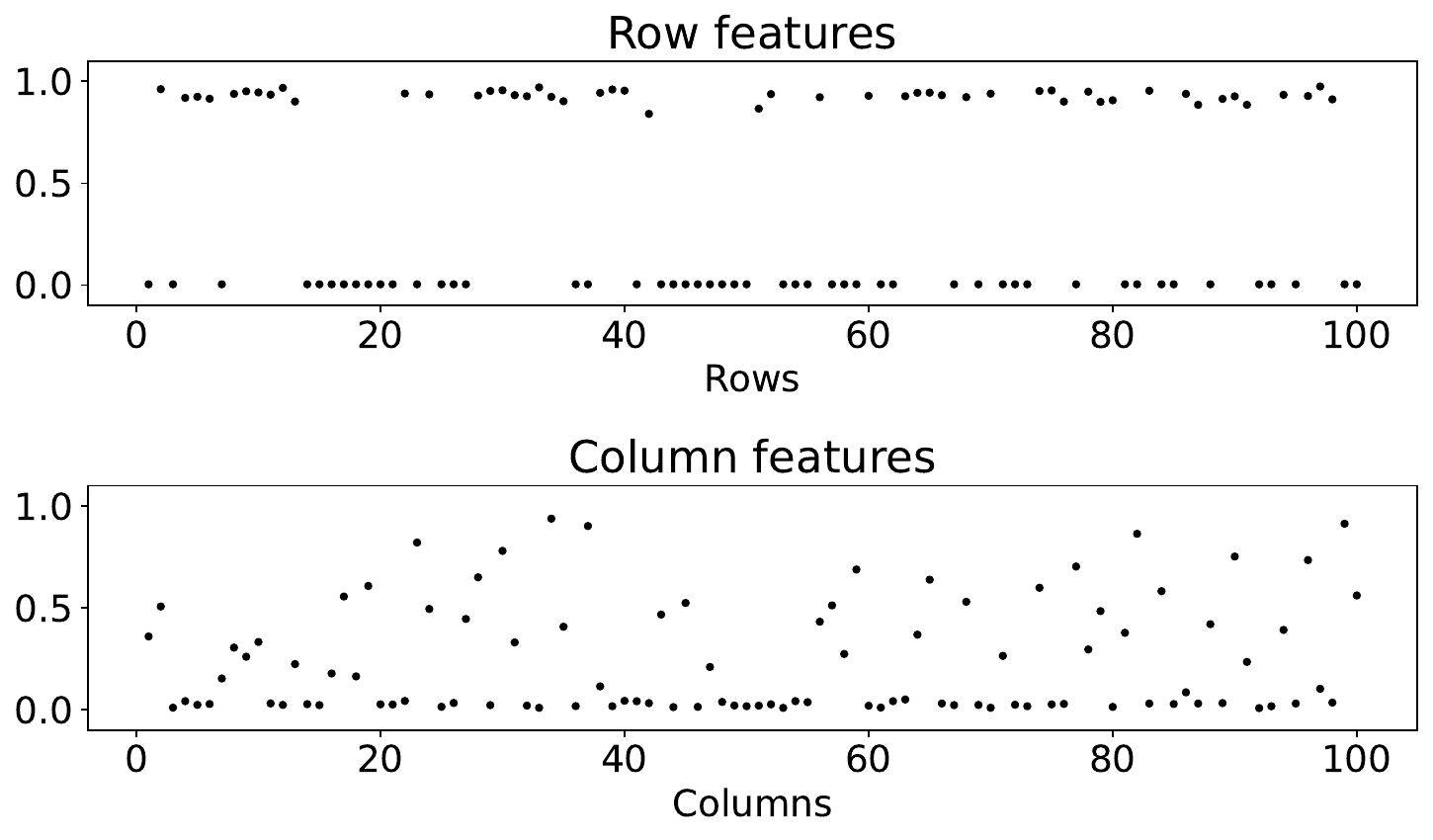}\hspace{10mm}
  \includegraphics[width=0.45\hsize]{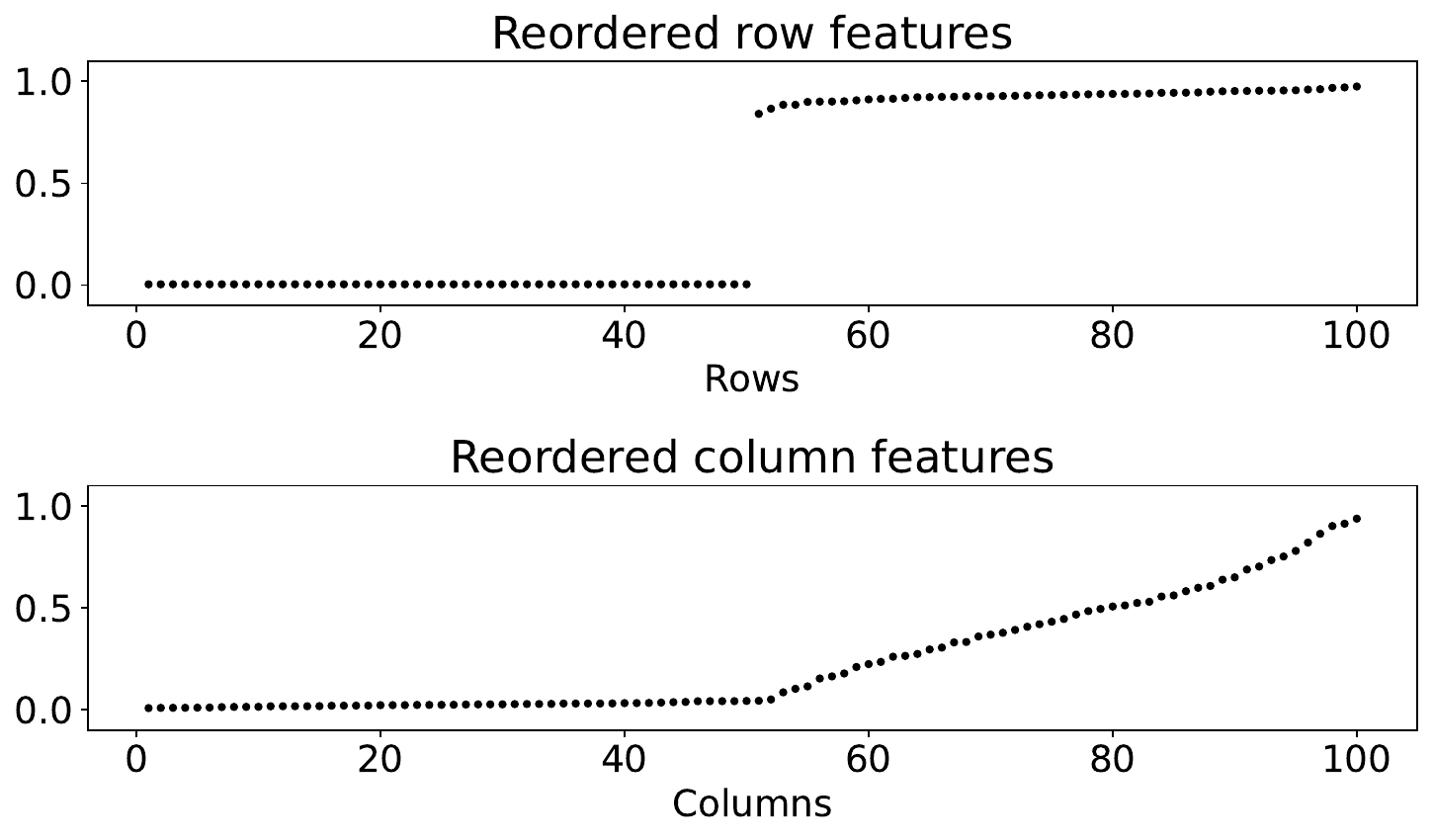}\vspace{-1mm}
  \caption{Results of the \textbf{gradation block model} for matrices $\bar{A}$, $A$, $\underline{A}$, $\underline{\hat{A}}$, and vectors $\bm{g}$, $\bm{h}$, $\underline{\bm{g}}$, $\underline{\bm{h}}$.}\vspace{3mm}
  \label{fig:results_m3}
\end{figure}

Figures \ref{fig:results_m1}, \ref{fig:results_m2}, and \ref{fig:results_m3} show the results of the LBM, SPM, and GBM, respectively. For each figure, we plotted matrix $\bar{A}$; observed matrix $A$; reordered observed and estimated matrices, $\underline{A}$ and $\underline{\hat{A}}$, respectively; row and column feature vectors, $\bm{g}$ and $\bm{h}$, respectively; and their reordered versions, $\underline{\bm{g}}$ and $\underline{\bm{h}}$. From the figures of reordered matrices $\underline{A}$ and $\underline{\hat{A}}$, we see that the DeepTMR can successfully extract the latent row-column structures (i.e., block structure, striped pattern, and gradation block structure) of given observed matrices. Particularly, the figures of the reordered output matrices, $\underline{\hat{A}}$, show that the outputs of the DeepTMR network reflect the global structures of the given observed matrices. It must be noted that the order of the row and column indices in matrices $\underline{A}$ and $\underline{\hat{A}}$ that represents the latent structure is not always unique and, thus, it is not necessarily identical with that of original matrix $A$, as shown in these figures. For instance, the latent structures of the three models (i.e., LBM, SPM, and GBM) can also be represented by flipping or reversing the order of the row or column indices. Moreover, for an LBM, the arbitrary orders of the row or column clusters are permitted for representing the latent block structure. 

From the figures of vectors $\underline{\bm{g}}$ and $\underline{\bm{h}}$, we see they capture the one-dimensional features of each row and column. For example, in the LBM case in Figure \ref{fig:results_m1}, the row (column) feature values in the same row (column) cluster are more similar than those in the mutually different row (column) clusters. In the GBM case in Figure \ref{fig:results_m3}, the row features are divided into two groups: one which contains the gradation pattern block and the other which does not. As for the column features, their values increase continuously within the gradation pattern block, whereas the remaining feature values are almost constant. 



\subsection{Comparison with existing matrix reordering methods}
\label{sec:comparison}

We also conducted a quantitative comparison between DeepTMR and the existing matrix reordering methods introduced in Section \ref{sec:related}. For comparison, we chose the spectral/dimension-reduction methods based on SVD-Rank-One, SVD-Angle, and MDS, whose algorithms are described in Section \ref{sec:related}. For the quantitative evaluation of these methods, we generated synthetic data matrices with true row/column orders, applied proposed and conventional methods, and compared their accuracies in matrix reordering. For simplicity, we considered the following data matrices, whose true row/column orders can be represented uniquely, except for row/column flipping. 

\paragraph{Diagonal gradation model} We generated a matrix $\bar{A}^{(0)} \in \mathbb{R}^{n \times p}$ with the following diagonal gradation pattern, setting the matrix size at $(n, p) = (100, 100)$. We assumed that the $(i, j)$th entry $\bar{A}^{(0)}_{ij}$ of matrix $\bar{A}^{(0)}$ is generated from a Gaussian distribution with mean $B_{ij}$ given by: 
\begin{align}
&B_{ij} = 0.9 - 0.8 \frac{|i - j|}{\max \{ n, p \}},\ \ \ 
i = 1, \dots, n, \ \ \ j = 1, \dots, p. 
\label{eq:B_ij_syn4}
\end{align}
For the standard deviation, we tried the following $10$ settings: $\sigma_t = 0.03t$ for $t = 1, \dots, 10$. As in Section \ref{sec:preliminary_exp}, we defined matrix $\bar{A}$ using matrix $\bar{A}^{(0)}$ based on (\ref{eq:A_normalization}), and applied a random permutation to the rows and columns of matrix $\bar{A}$ to obtain observed matrix $A$. For each setting of $t$, we generated $10$ observed matrices and applied the DeepTMR, SVD-Rank-One, SVD-Angle, and MDS. Because the training result of the DeepTMR depends on its initial parameters and the selection of the mini-batch for each iteration, for the same observed matrix, $A$, we trained the DeepTMR model five times and adopted the trained model with the minimum mean training loss for the last  $100$ iterations. The other hyperparameter settings for training the DeepTMR are listed in Table \ref{tab:hyperparameter}. 

To quantitatively evaluate these methods, we computed the following matrix reordering error. Let $P \in \mathbb{R}^{n \times p}$ and $\bar{P} \in \mathbb{R}^{n \times p}$ be the mean matrices, whose $(i, j)$th entry is the (population) mean of $A_{ij}$ and $\bar{A}_{ij}$, respectively. Let $\pi^{\mathrm{row} (0)}$ and $\pi^{\mathrm{column} (0)}$, respectively be the permutations of $\{ 1, 2, \dots, n \}$ and $\{ 1, 2, \dots, p \}$, which indicate the order of the rows and columns determined by each method. The flipped versions of these orders are defined as $\pi^{\mathrm{row} (1)}$ and $\pi^{\mathrm{column} (1)}$ (i.e., $\pi^{\mathrm{row} (0)} (i) = \pi^{\mathrm{row} (1)} (n - i + 1)$ and $\pi^{\mathrm{column} (0)} (j) = \pi^{\mathrm{column} (1)} (p - j + 1)$ for $i = 1, \dots, n$ and $j = 1, \dots, p$). 
Let $\bar{\pi}^{\mathrm{row}}$ and $\bar{\pi}^{\mathrm{column}}$, respectively be the order of the rows and columns that reconstruct original (correctly ordered) matrix $\bar{A}$. Both $\pi^{\mathrm{row} (0)} = \bar{\pi}^{\mathrm{row}}$ and $\pi^{\mathrm{row} (1)} = \bar{\pi}^{\mathrm{row}}$ indicate that the correct row ordering is obtained. Based on this fact, we redefine the row/column orders, $\pi^{\mathrm{row}}$ and $\pi^{\mathrm{column}}$, obtained by each method as follows: 
\begin{align}
&(\hat{k}, \hat{h}) = \argmin_{(k, h) \in \{(0, 0), (0, 1), (1, 0), (1, 1) \}} \frac{1}{np} \sum_{i = 1}^n \sum_{j = 1}^p \left( \bar{P}_{ij} - P_{\pi^{\mathrm{row} (k)} (i) \pi^{\mathrm{column} (h)} (j)} \right)^2, \nonumber \\
&\pi^{\mathrm{row}} = \pi^{\mathrm{row} (\hat{k})}, \ \ \ \ \ 
\pi^{\mathrm{column}} = \pi^{\mathrm{column} (\hat{h})}. 
\end{align}
Finally, we define the matrix reordering error $E$ as: 
\begin{align}
E = \frac{1}{np} \sum_{i = 1}^n \sum_{j = 1}^p \left( \bar{P}_{ij} - P_{\pi^{\mathrm{row}} (i) \pi^{\mathrm{column}} (j)} \right)^2. 
\end{align}

\begin{figure}[p]
  \centering
  \includegraphics[width=0.92\hsize]{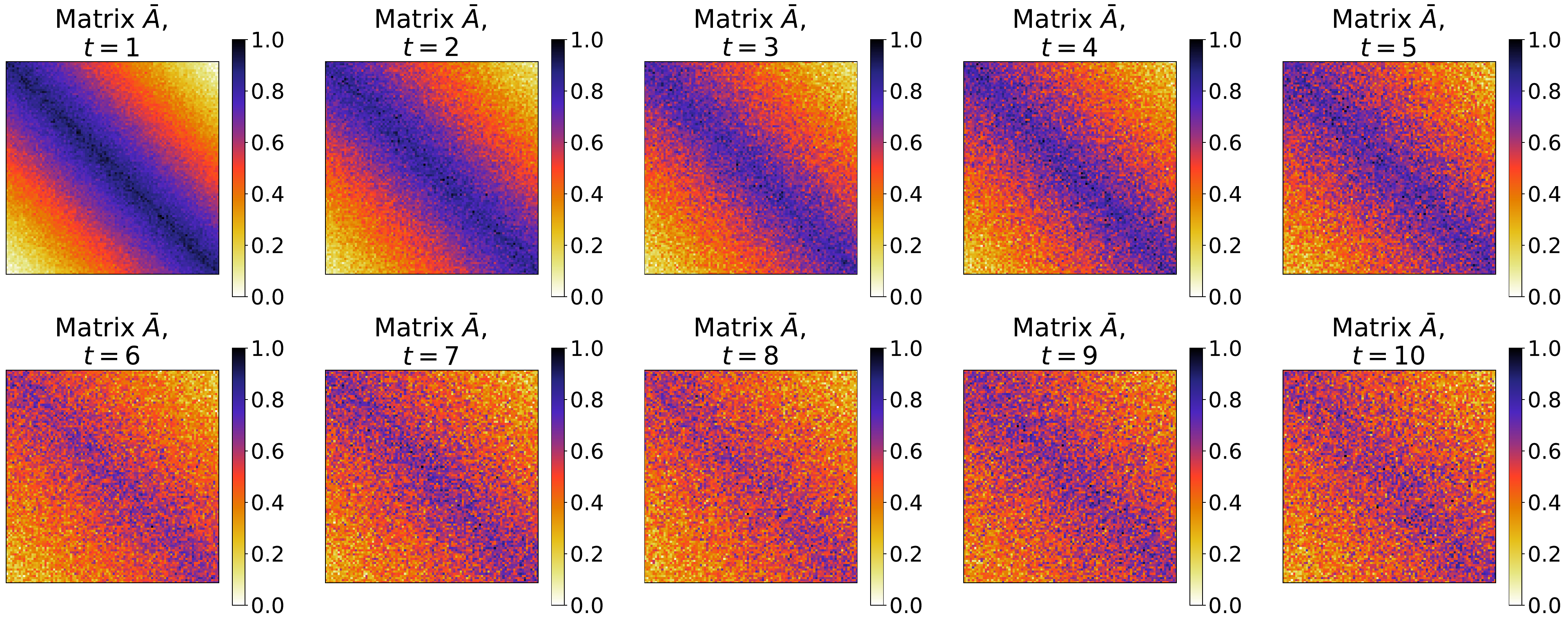}
  \caption{Examples of matrix $\bar{A}$ for the \textbf{diagonal gradation model} with different levels of noise standard deviation.}\vspace{3mm}
  \label{fig:syn4_A_bar}
  \includegraphics[width=0.92\hsize]{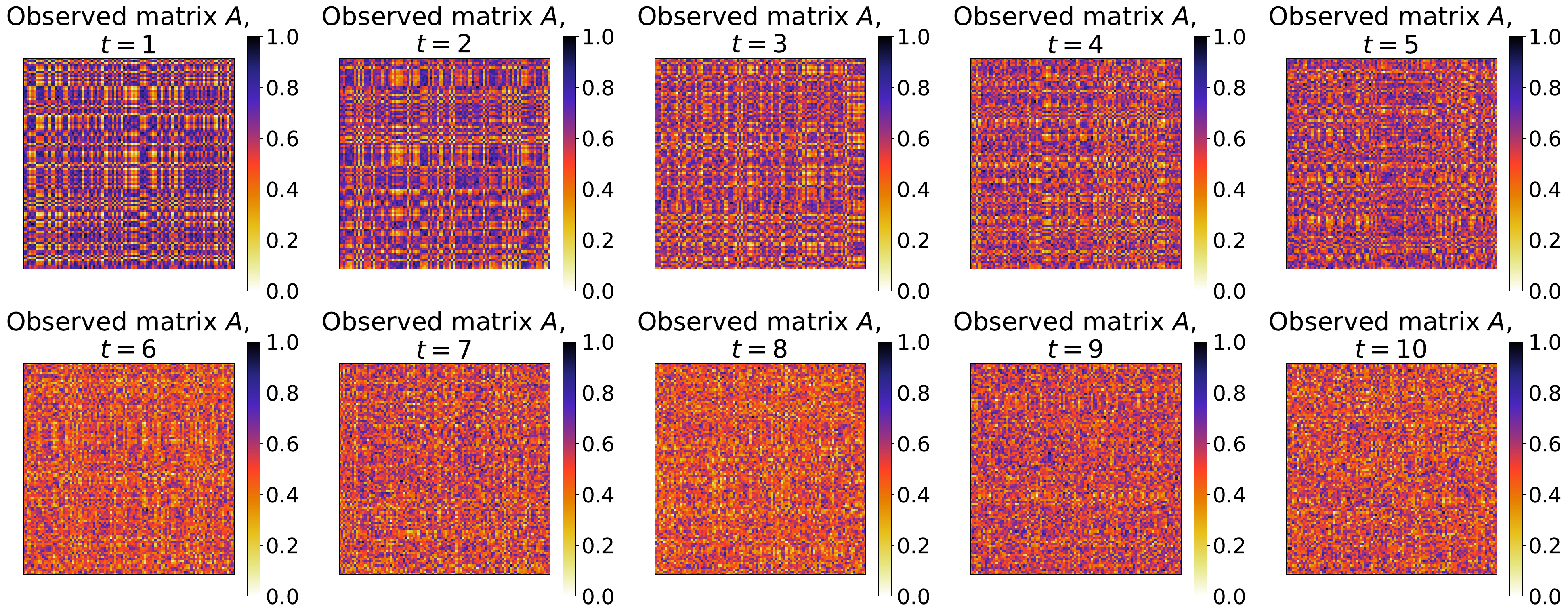}
  \caption{Examples of observed matrix $A$ for the \textbf{diagonal gradation model} with different levels of noise standard deviation.}\vspace{3mm}
  \label{fig:syn4_A}
  \includegraphics[width=0.92\hsize]{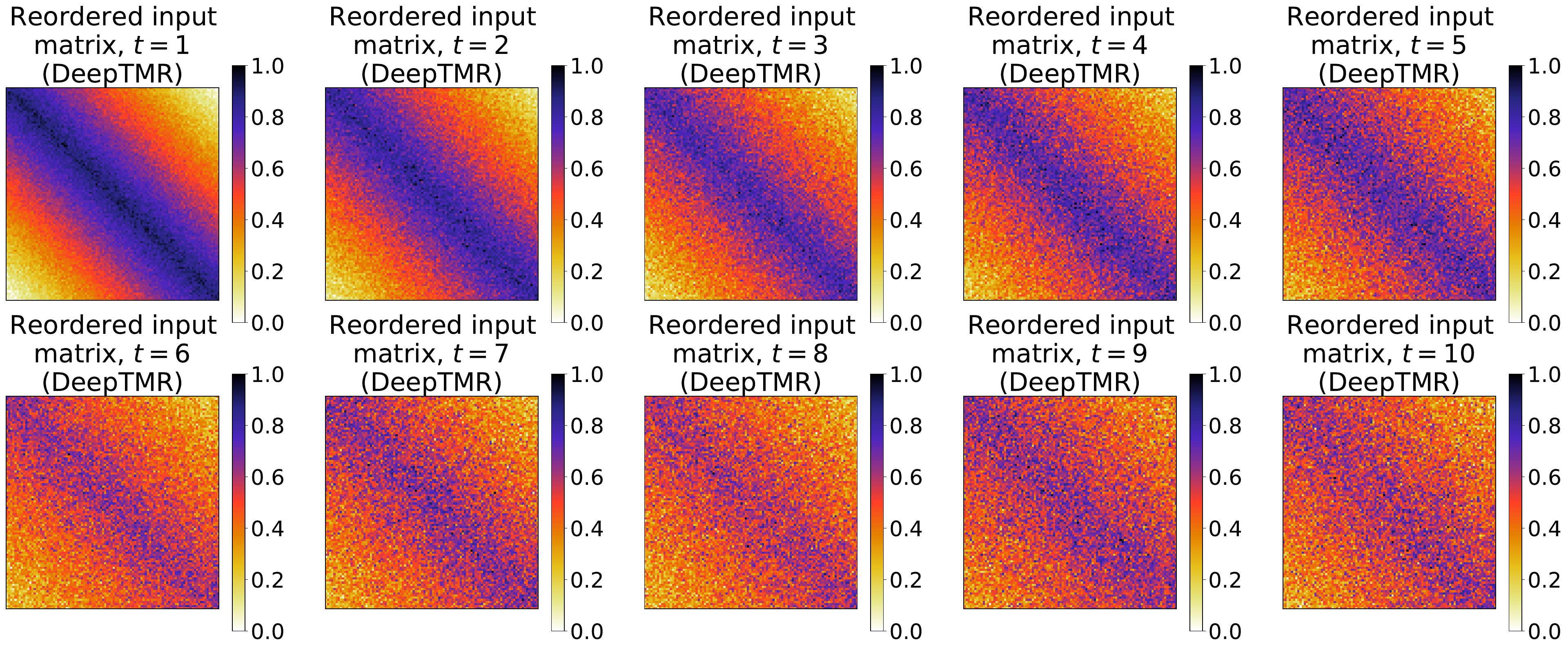}
  \caption{Examples of reordered input matrix $\underline{A}$ for the \textbf{diagonal gradation model} with different levels of noise standard deviation (\textbf{DeepTMR}).}
  \label{fig:syn4_A_underline}
\end{figure}

\begin{figure}[p]
  \centering
  \includegraphics[width=0.92\hsize]{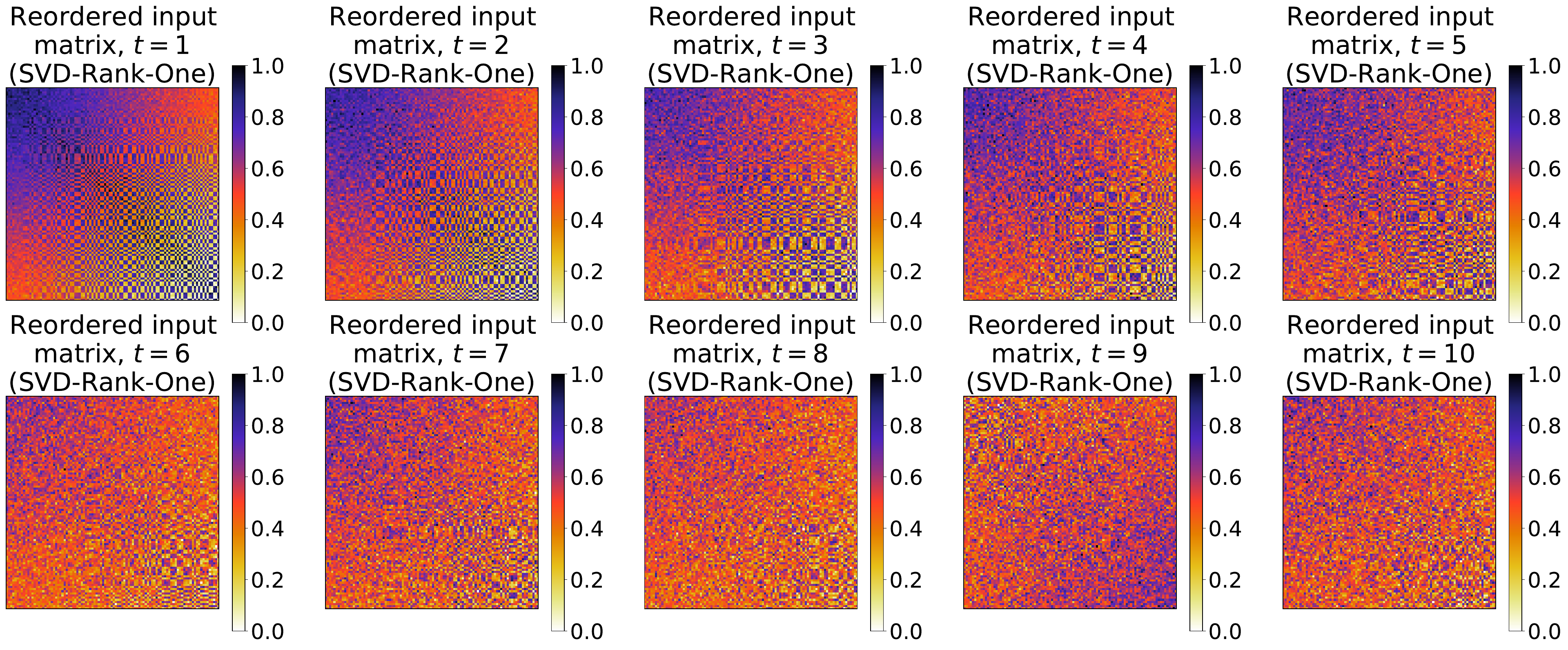}
  \caption{Examples of reordered input matrix $\underline{A}$ for the \textbf{diagonal gradation model} with different levels of noise standard deviation (\textbf{SVD-Rank-One}).}\vspace{3mm}
  \label{fig:syn4_A_underline_SVD}
  \includegraphics[width=0.92\hsize]{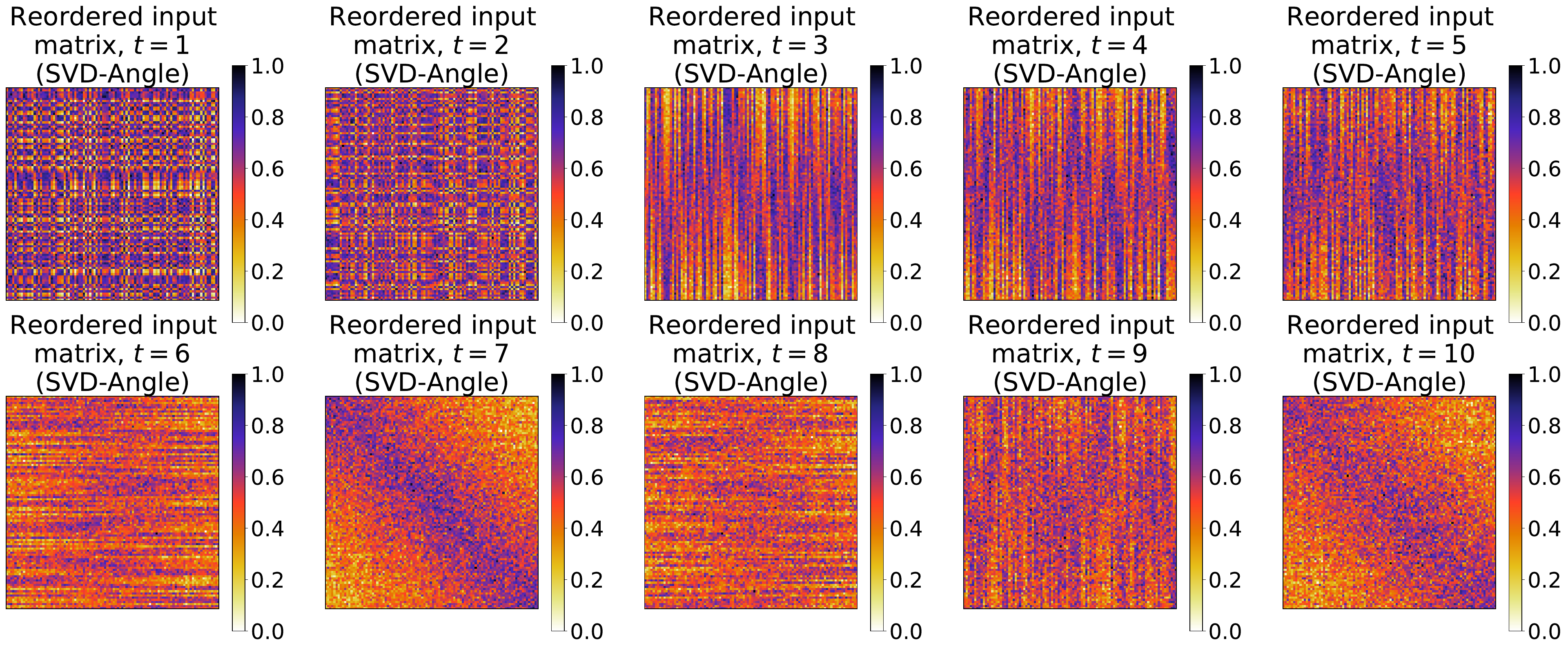}
  \caption{Examples of reordered input matrix $\underline{A}$ for the \textbf{diagonal gradation model} with different levels of noise standard deviation (\textbf{SVD-Angle}).}\vspace{3mm}
  \label{fig:syn4_A_underline_PCA}
  \includegraphics[width=0.92\hsize]{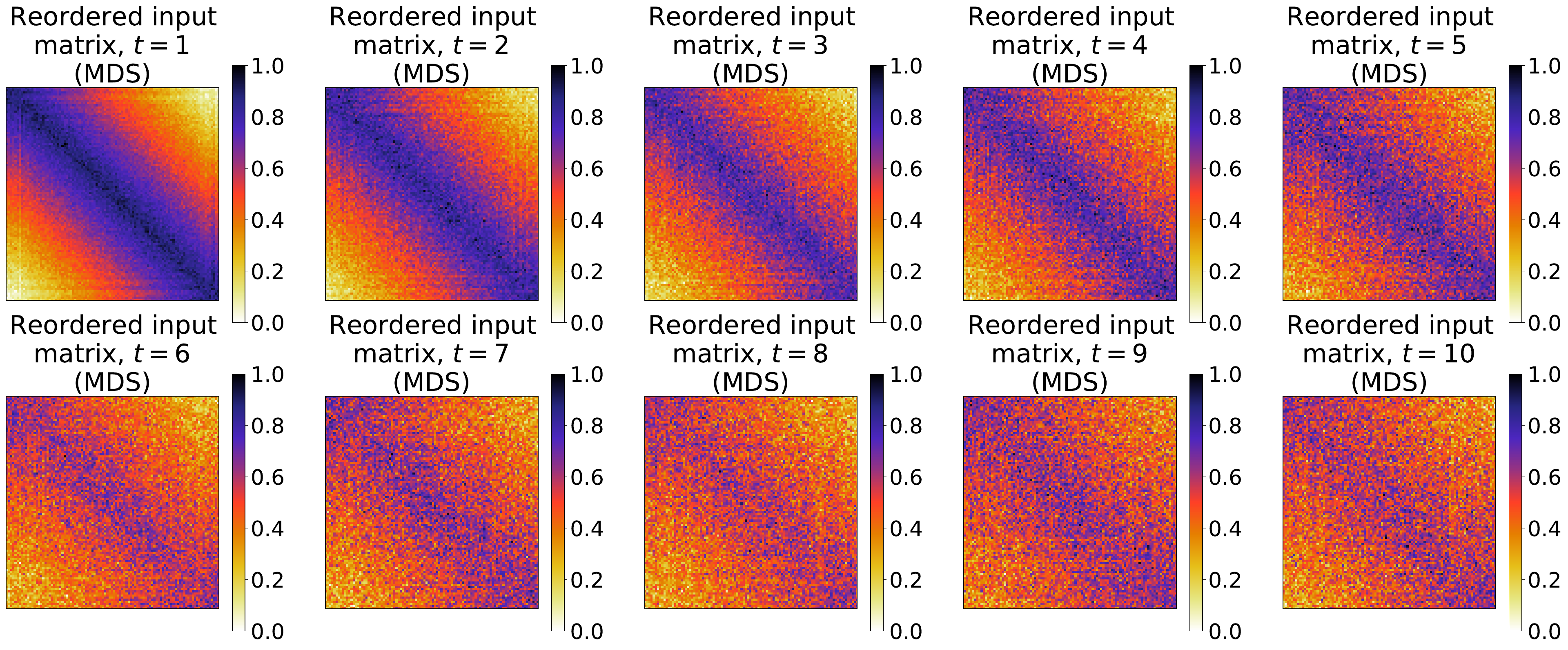}
  \caption{Examples of reordered input matrix $\underline{A}$ for the \textbf{diagonal gradation model} with different levels of noise standard deviation (\textbf{MDS}).}
  \label{fig:syn4_A_underline_MDS}
\end{figure}

\begin{figure}[t]
  \centering
  \includegraphics[width=0.92\hsize]{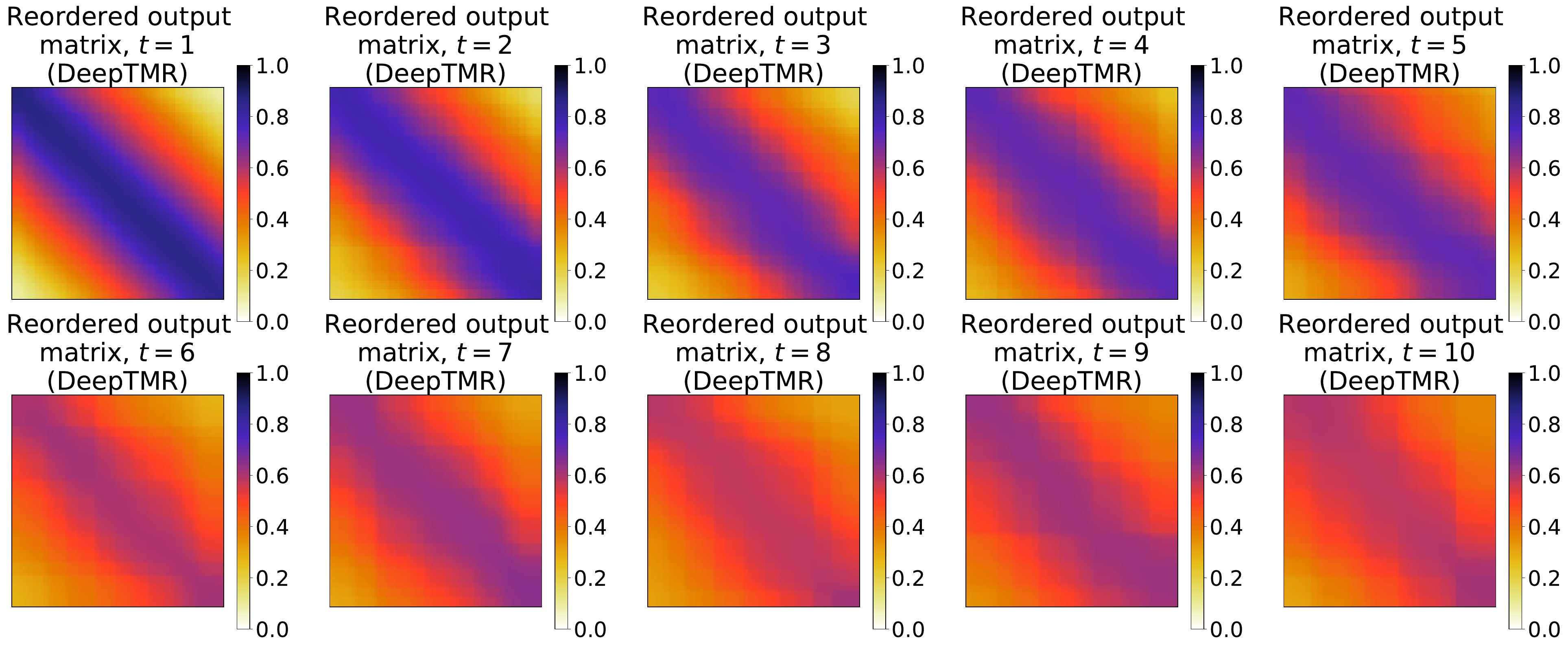}
  \caption{Examples of reordered output matrix $\underline{\hat{A}}$ for the \textbf{diagonal gradation model} with different levels of noise standard deviation (\textbf{DeepTMR}).}\vspace{3mm}
  \label{fig:results_m4_Aout}
  \includegraphics[width=0.65\hsize]{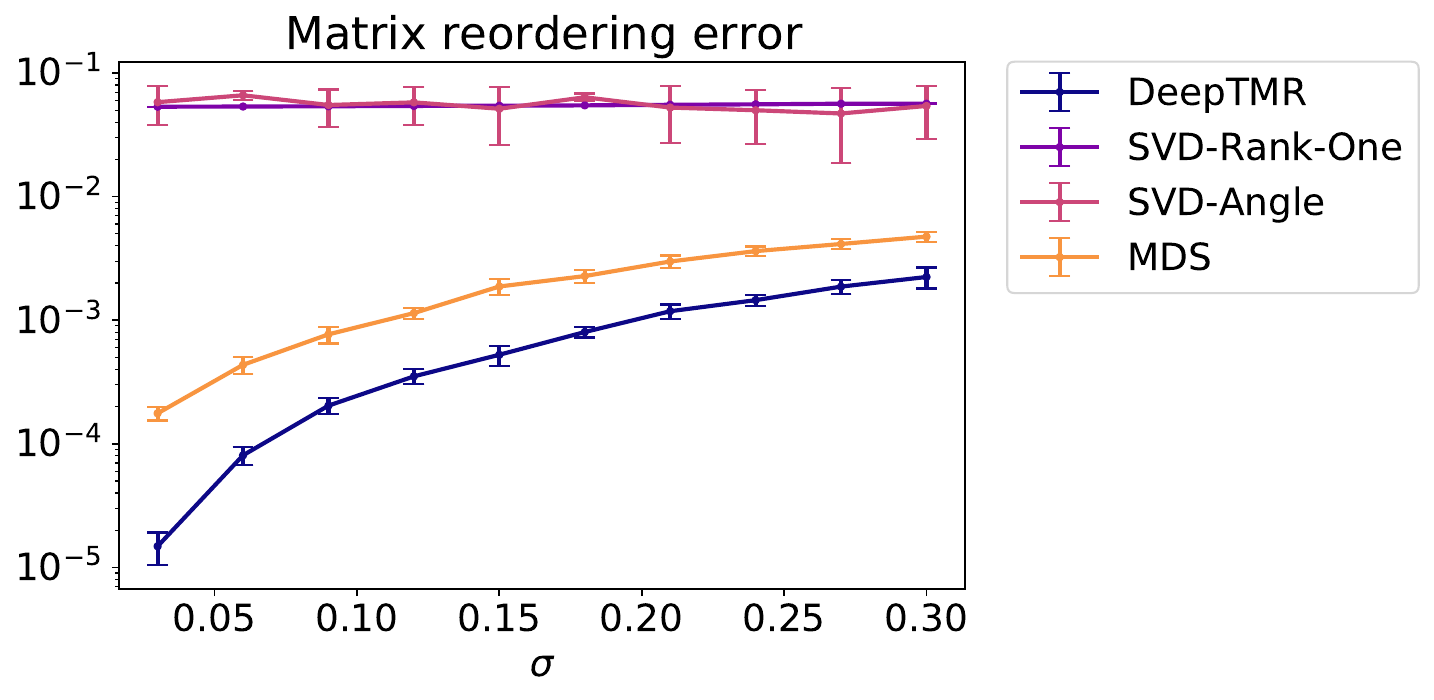}
  \caption{Matrix reordering error of the DeepTMR, SVD-Rank-One, SVD-Angle, and MDS. The error bars indicate the sample standard deviations of the results for $10$ trials.}
  \label{fig:results_m4_compare}
\end{figure}

Figures \ref{fig:syn4_A_bar} and \ref{fig:syn4_A} respectively show the examples of matrices $\bar{A}$ and $A$ with different levels of noise standard deviation $\sigma_t$, where $t = 1, \dots, 10$. 
Figures \ref{fig:syn4_A_underline}, \ref{fig:syn4_A_underline_SVD}, \ref{fig:syn4_A_underline_PCA}, and \ref{fig:syn4_A_underline_MDS} respectively show the examples of the reordered observed matrix $\underline{A}$ based on row/column orderings $(\pi^{\mathrm{row}}, \pi^{\mathrm{column}})$ obtained by DeepTMR, SVD-Rank-One, SVD-Angle, and MDS. Figure \ref{fig:results_m4_Aout} shows the reordered output matrix $\underline{\hat{A}}$ for the DeepTMR. From these figures, the DeepTMR and MDS can relatively successfully reorder the observed matrix compared to the SVD-based methods. 
Figure \ref{fig:results_m4_compare} shows the matrix reordering error of the DeepTMR, SVD-Rank-One, SVD-Angle, and MDS. This figure shows that the DeepTMR can achieve the minimum matrix reordering error in this setting compared to the other three methods. 



\subsection{Experiment using the divorce predictors dataset}
\label{sec:divorce_data}

Next, we applied the DeepTMR to the divorce predictors dataset \cite{Yontem2017, Yontem2019} from the UCI Machine Learning Repository \cite{Dua2017}. The original data matrix, $A^{(0)}$, consists of $170$ rows and $54$ columns, which represent the questionnaire respondents and their attributes, respectively. Each entry $A^{(0)}_{ij} \in \{ 0, 1, \dots, 4 \}$ shows the Divorce Predictors Scale (DPS), with a higher value indicating a higher divorce risk. The meaning of the five-factor scale is as follows: $0$ for ``Never,'' $1$ for ``Rarely,'' $2$ for ``Occasionally,'' $3$ for ``Often,'' and $4$ for ``Always,'' for Attributes $31$ to $54$, while they are reversed (i.e., $0$ for ``Always'' and $4$ for ``Never'') for Attributes $1$ to $30$. The meaning of each attribute index of this dataset is provided in Appendix \ref{sec:ap_table_divorce}. 

As in Section \ref{sec:preliminary_exp}, we defined observed matrix $A$ based on (\ref{eq:A_normalization}) by replacing $\bar{A}^{(0)}$ and $\bar{A}$ with $A^{(0)}$ and $A$, respectively. Then, we applied DeepTMR to observed matrix $A$ and checked the latent row-column structure of matrix $A$ extracted by the DeepTMR. The hyperparameter settings for training the DeepTMR are listed in Table \ref{tab:hyperparameter}. 

\begin{figure}[t]
  \centering
  \includegraphics[width=0.495\hsize]{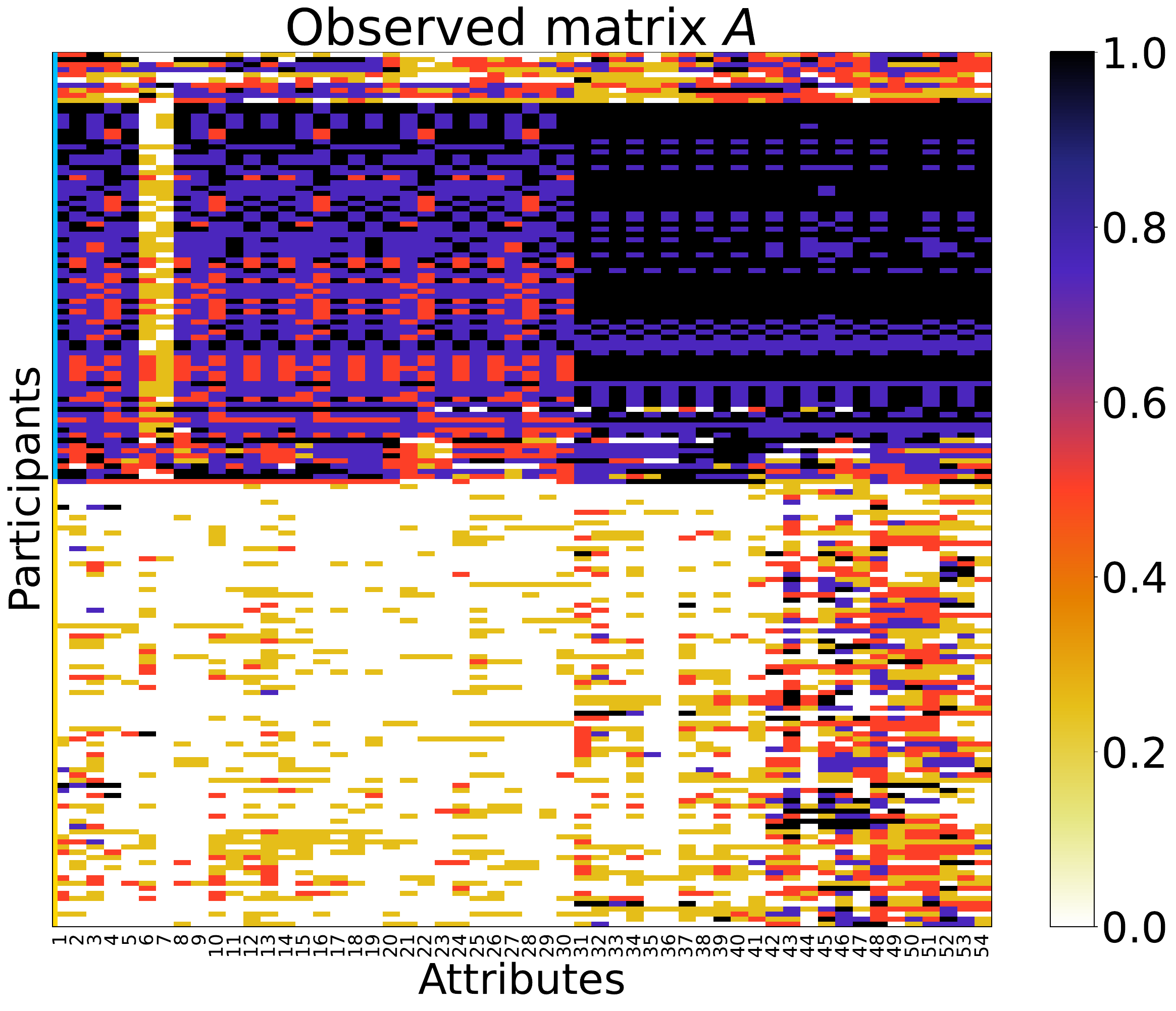}
  \includegraphics[width=0.495\hsize]{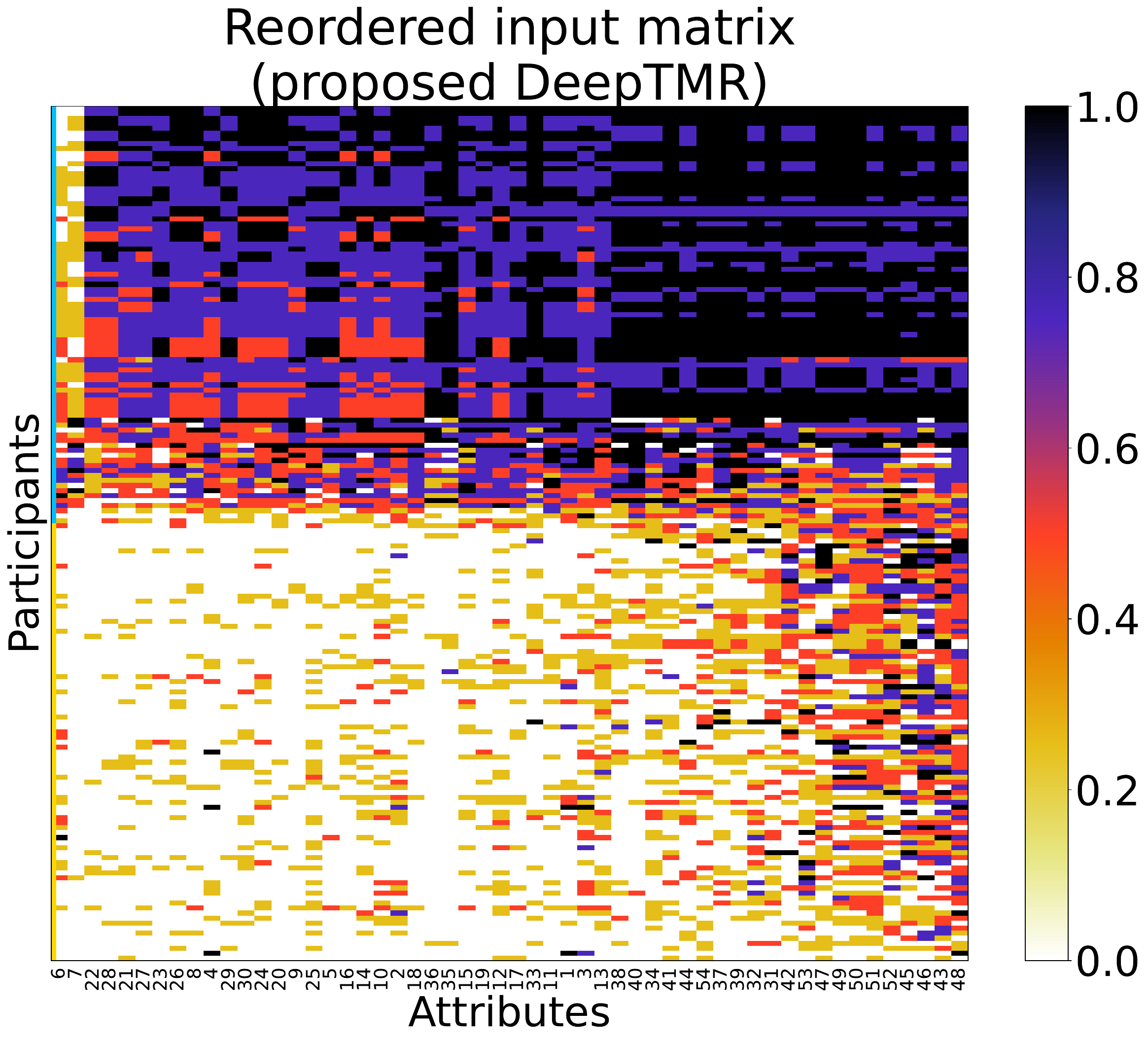}\\
  \includegraphics[width=0.495\hsize]{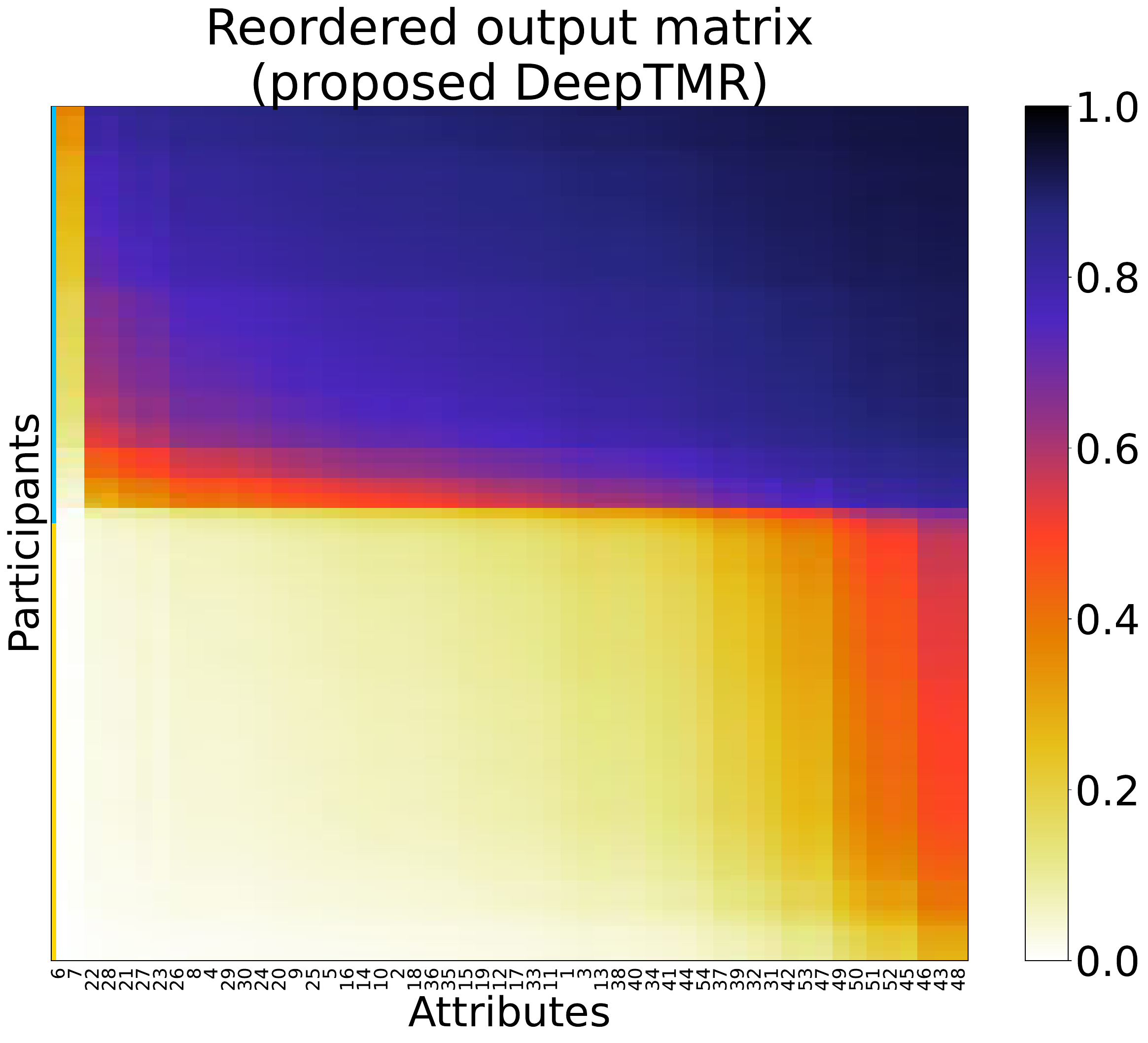}
  \includegraphics[width=0.2\hsize]{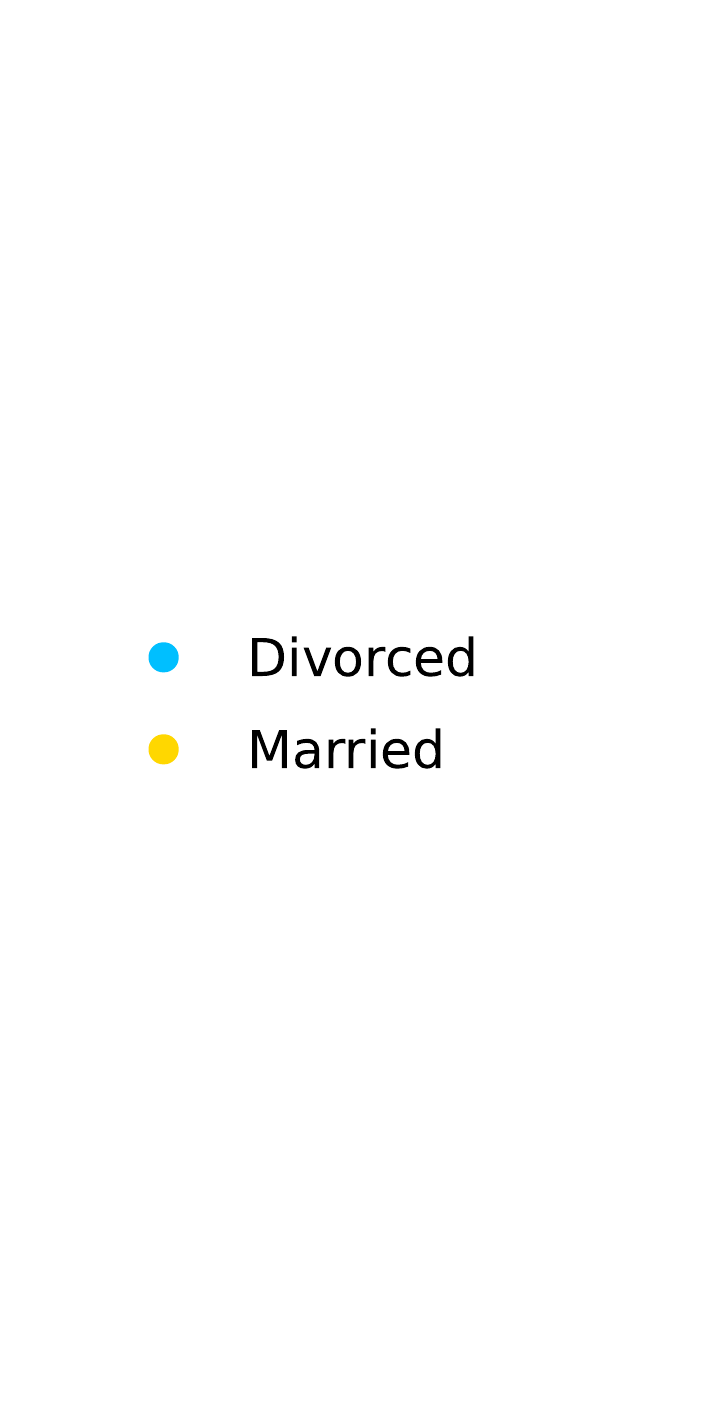}\\
  \includegraphics[width=0.45\hsize]{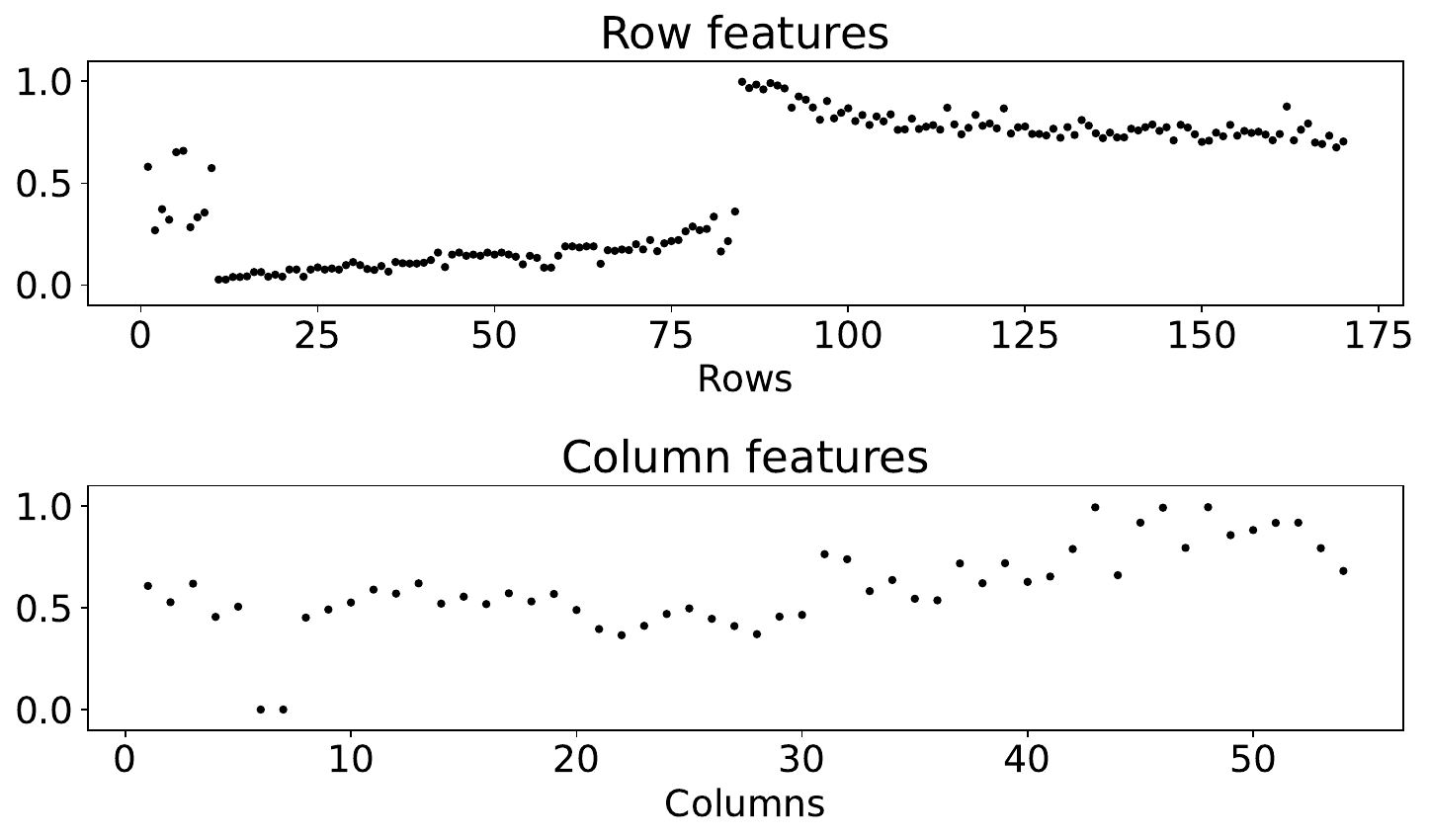}\hspace{10mm}
  \includegraphics[width=0.45\hsize]{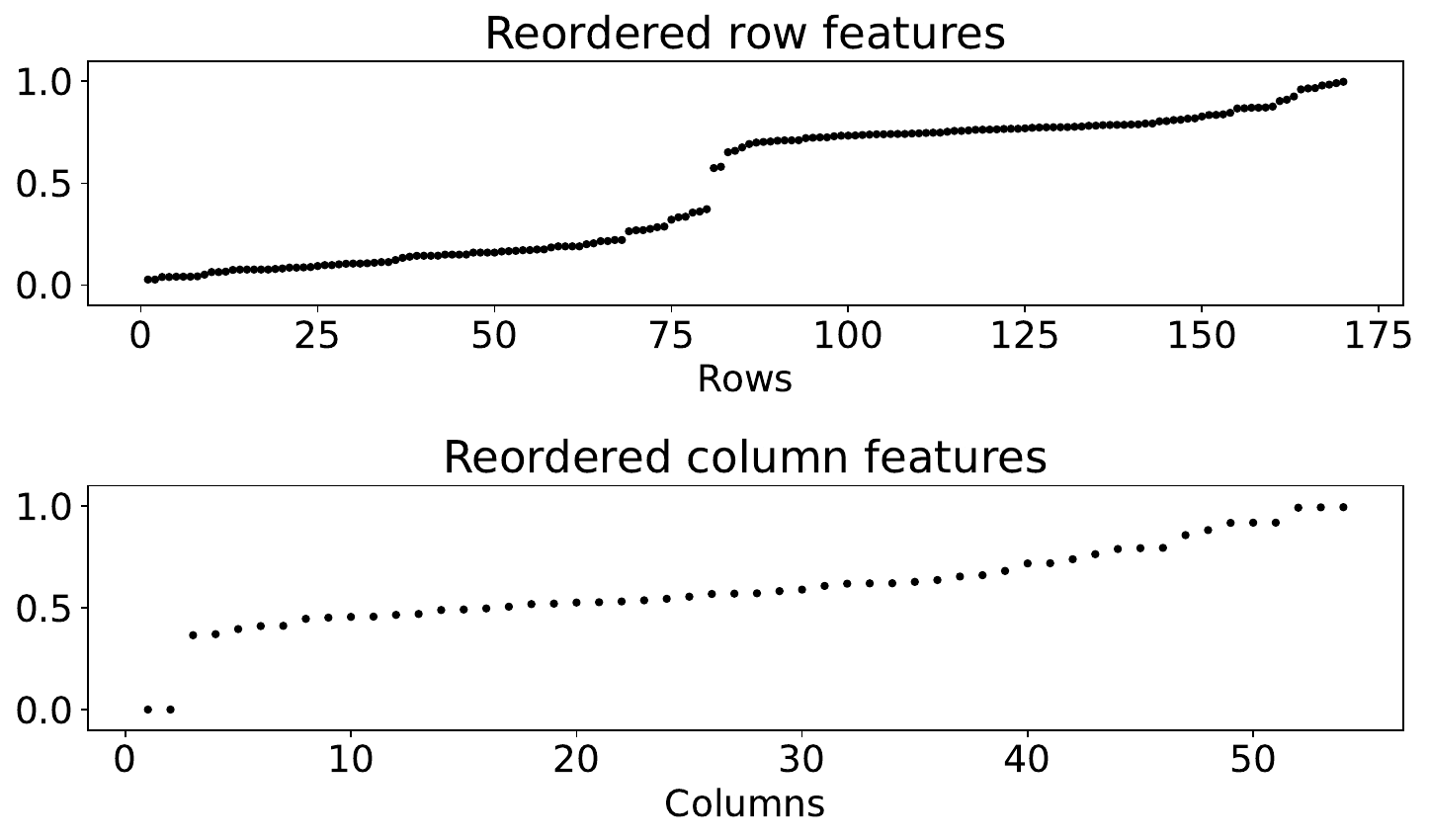}\vspace{-1mm}
  \caption{Results of the \textbf{divorce predictors dataset} for matrices $A$, $\underline{A}$, $\underline{\hat{A}}$, and vectors $\bm{g}$, $\bm{h}$, $\underline{\bm{g}}$, $\underline{\bm{h}}$.}\vspace{3mm}
  \label{fig:results_p1}
\end{figure}

Figure \ref{fig:results_p1} shows the results of matrices $A$, $\underline{A}$, $\underline{\hat{A}}$, and vectors $\bm{g}$, $\bm{h}$, $\underline{\bm{g}}$, $\underline{\bm{h}}$ for this dataset. For each row in matrices $A$, $\underline{A}$, and $\underline{\hat{A}}$, the class labels ``divorced'' or ``married'' are shown in different colors on the left-hand side of the matrix. From the reordered input and output matrices $\underline{A}$ and $\underline{\hat{A}}$, we respectively see the latent row-column structure of the observed matrix. Roughly, the DPS takes higher values in the ``divorced'' rows than in the ``married'' ones. However, some divorced participants show relatively low DPS for some attributes (e.g., Attributes $21$, $22$, and $28$). We also see that both the divorced and married participants show relatively high DPS for Attributes $43$ and $48$, whereas most participants show relatively low DPS for Attributes $6$ and $7$, both items referring to how to behave at home with a partner. 

\subsection{Experiment using the metropolis traffic census dataset}
\label{sec:traffic_data}

We also applied the DeepTMR to the metropolis traffic census dataset from e-Stat \cite{estat}. The rows and columns of the relational data matrix of this dataset represent the locations of metropolitan areas in Japan, each $(i, j)$th entry showing the number of people commuting (to work or school) one way from the $i$th location to the $j$th location per day. We removed the rows and columns that represent unknown locations (e.g., ``unknown below Tokyo'') and the total of multiple locations (e.g., ``total of three wards in central Tokyo'') from the original dataset. Let $A^{(0)} \in \mathbb{R}^{n \times p}$ be the matrix after removing the rows and columns, where $n = p = 249$. 
To alleviate the significant differences between entry values and consider relatively small entry values, we defined matrix $A^{(1)} \in \mathbb{R}^{n \times p}$, whose entries are given by $A^{(1)}_{ij} = \log \left(A^{(0)}_{ij} + 1 \right)$ for $i = 1, \dots, n$, and $j = 1, \dots, p$. 

As in Sections \ref{sec:preliminary_exp} and \ref{sec:divorce_data}, we defined observed matrix $A$ by replacing $\bar{A}^{(0)}$ and $\bar{A}$ with $A^{(1)}$ and $A$, respectively. Then, we applied DeepTMR to observed matrix $A$ and checked the latent row-column structure of matrix $A$ extracted by the DeepTMR. The hyperparameter settings for training the DeepTMR are listed in Table \ref{tab:hyperparameter}. 

\begin{figure}[p]
  \centering
  \includegraphics[width=0.45\hsize]{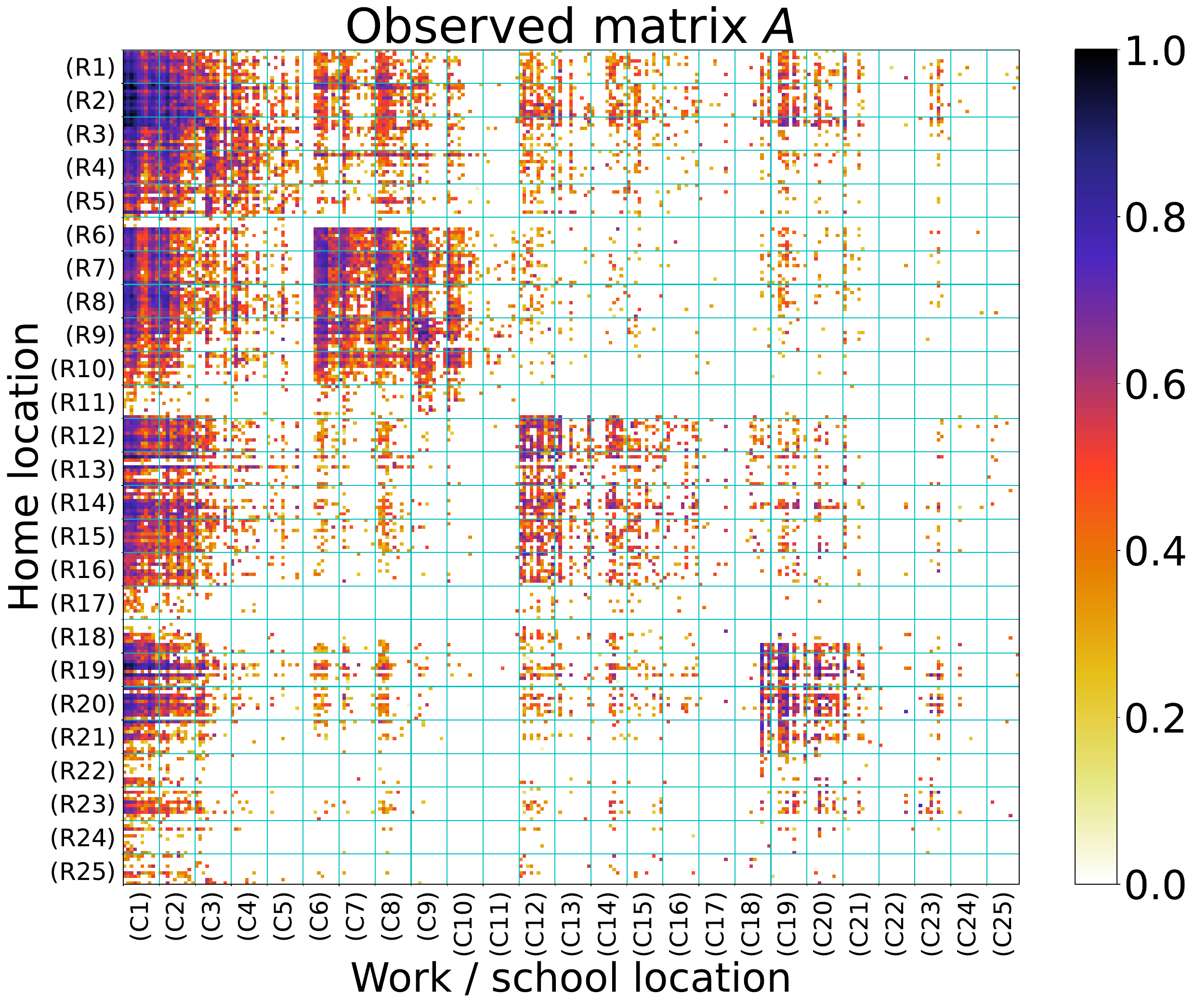}
  \includegraphics[width=0.25\hsize]{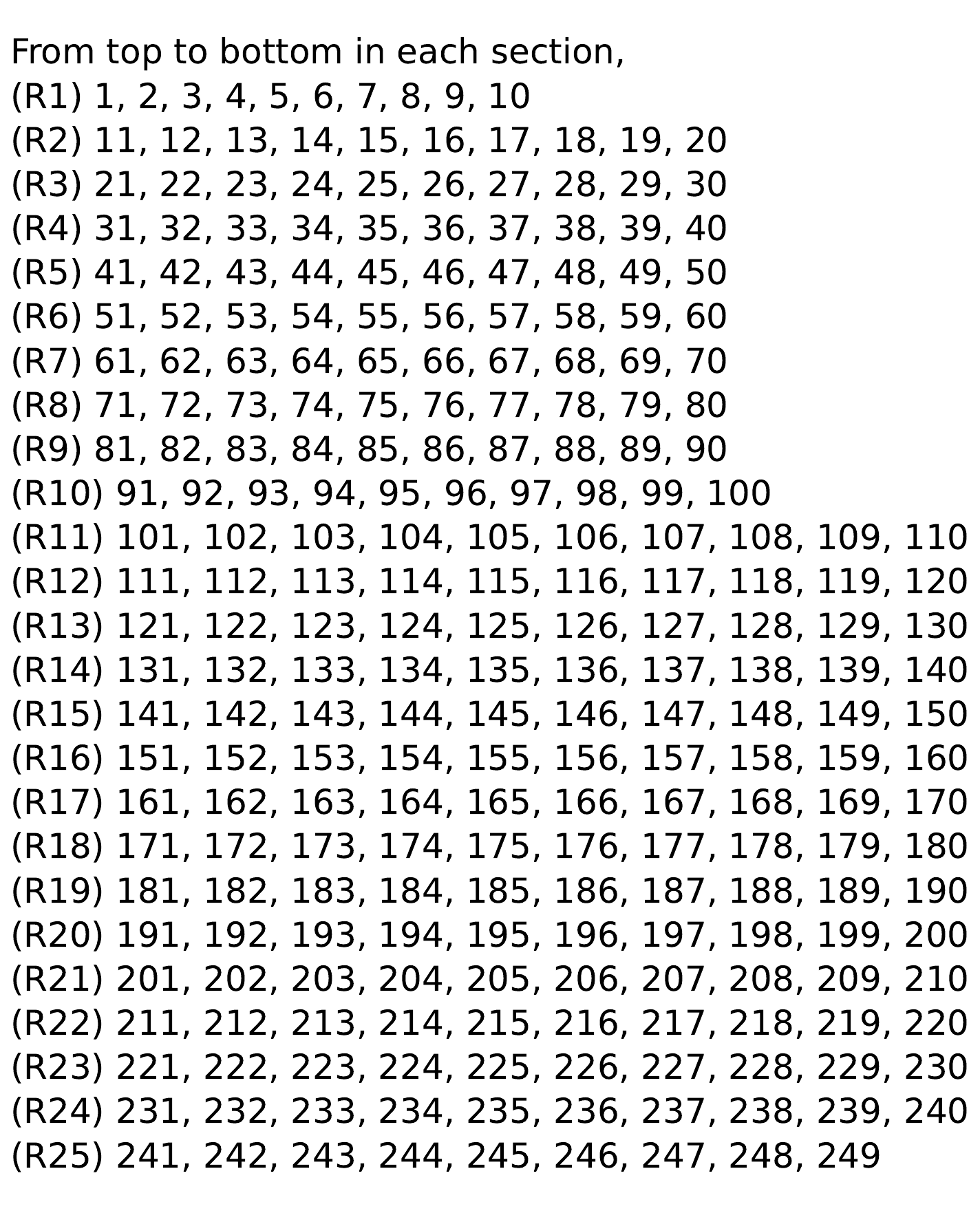}
  \includegraphics[width=0.25\hsize]{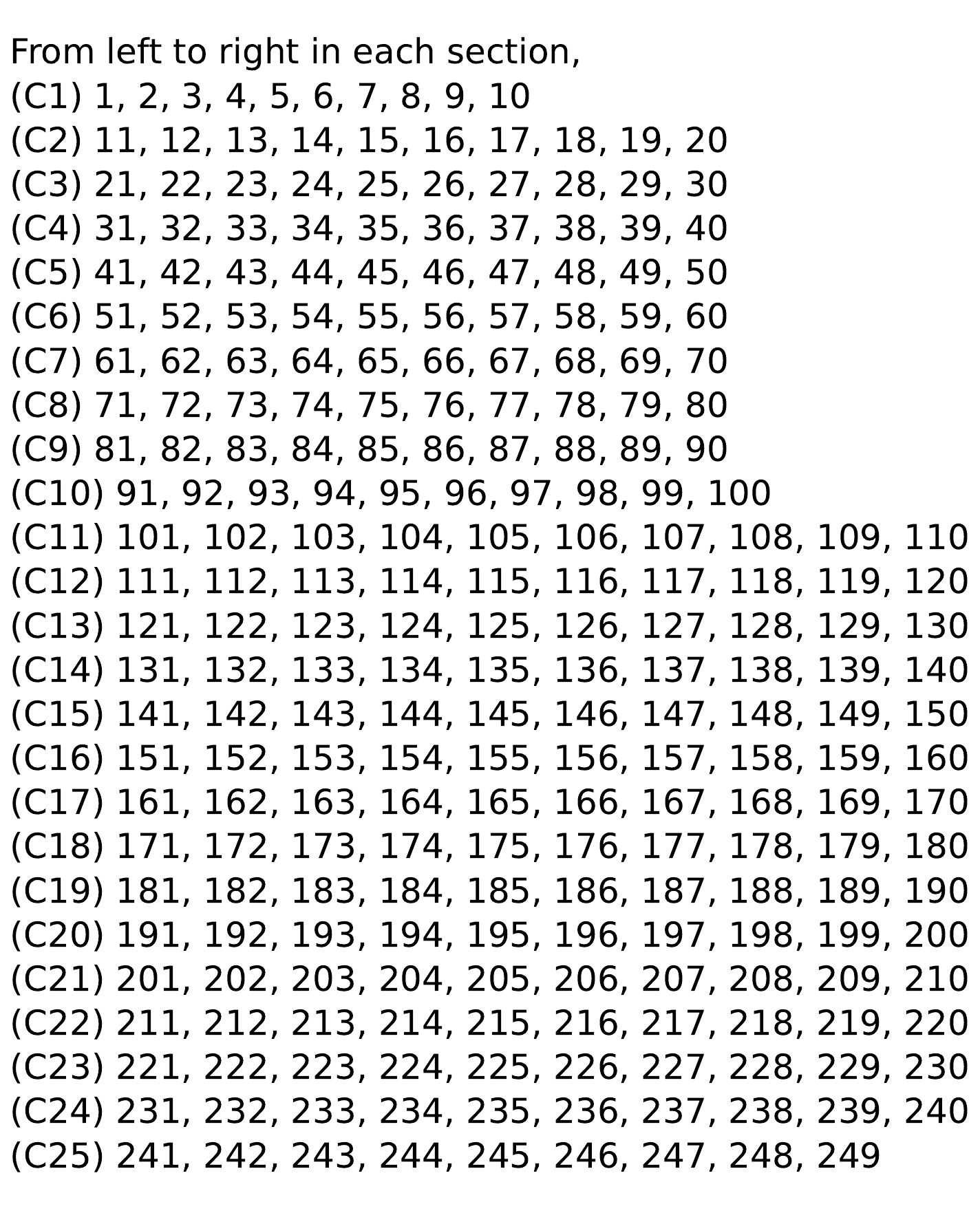}\vspace{3mm}
  \includegraphics[width=0.45\hsize]{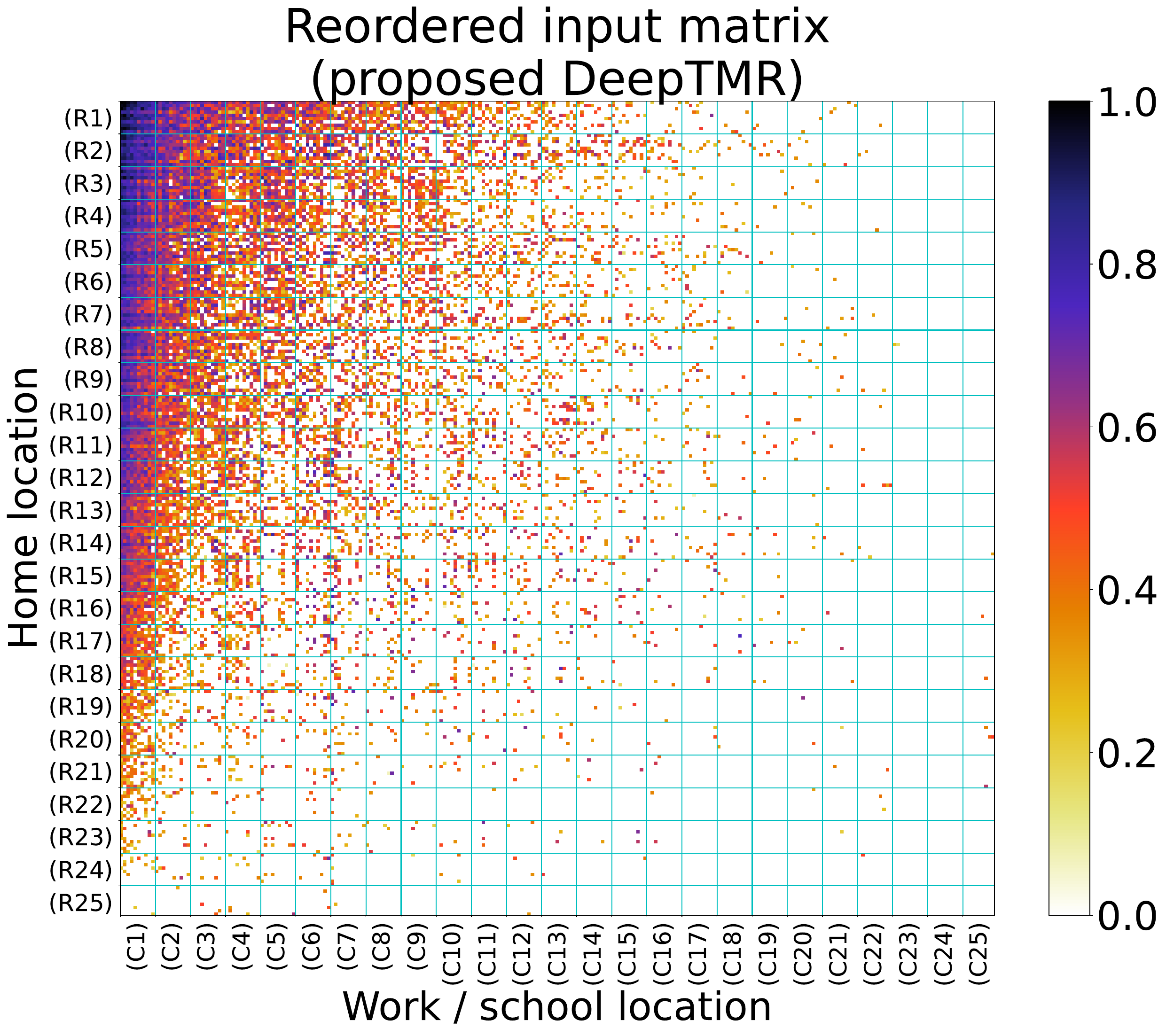}
  \includegraphics[width=0.25\hsize]{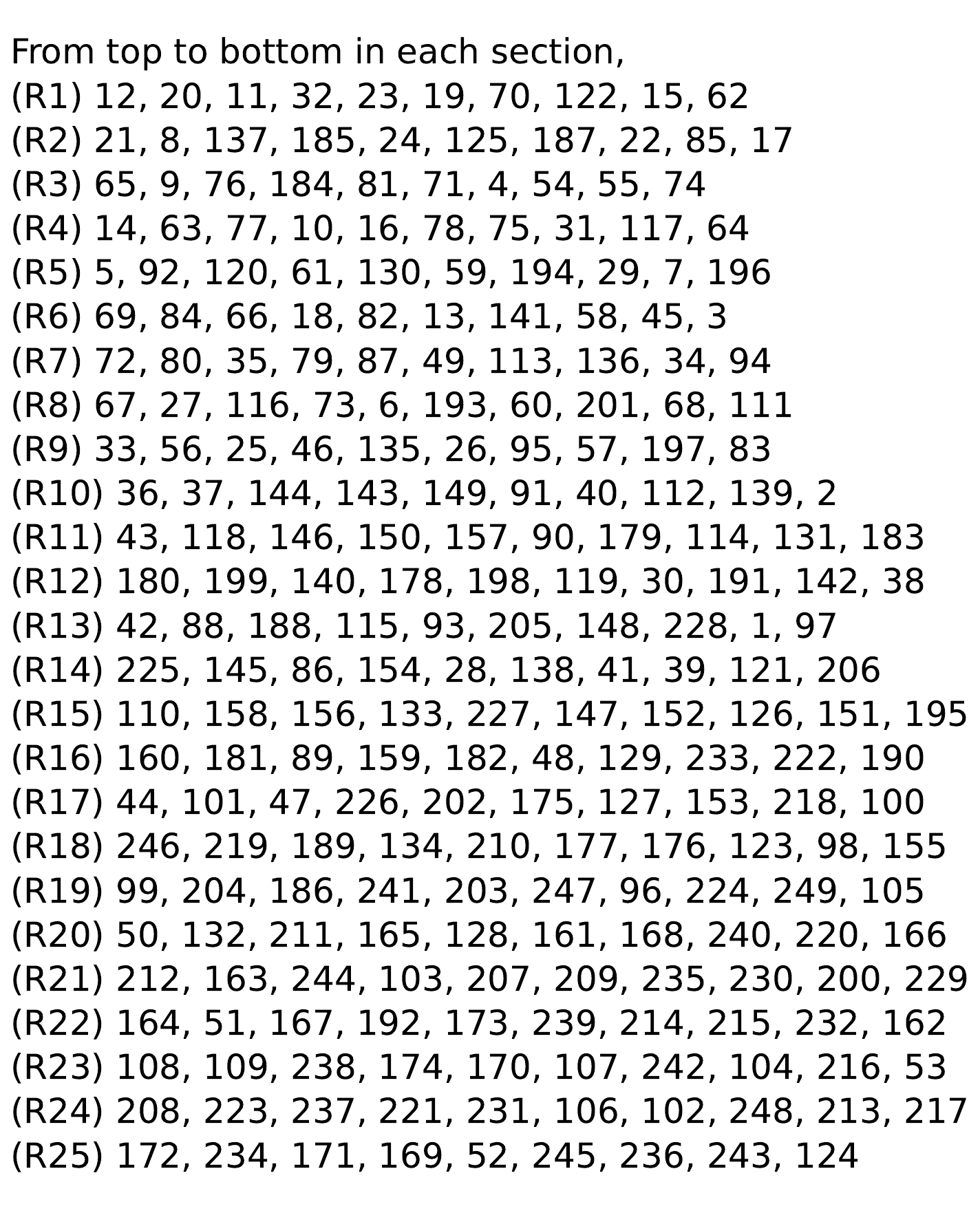}
  \includegraphics[width=0.25\hsize]{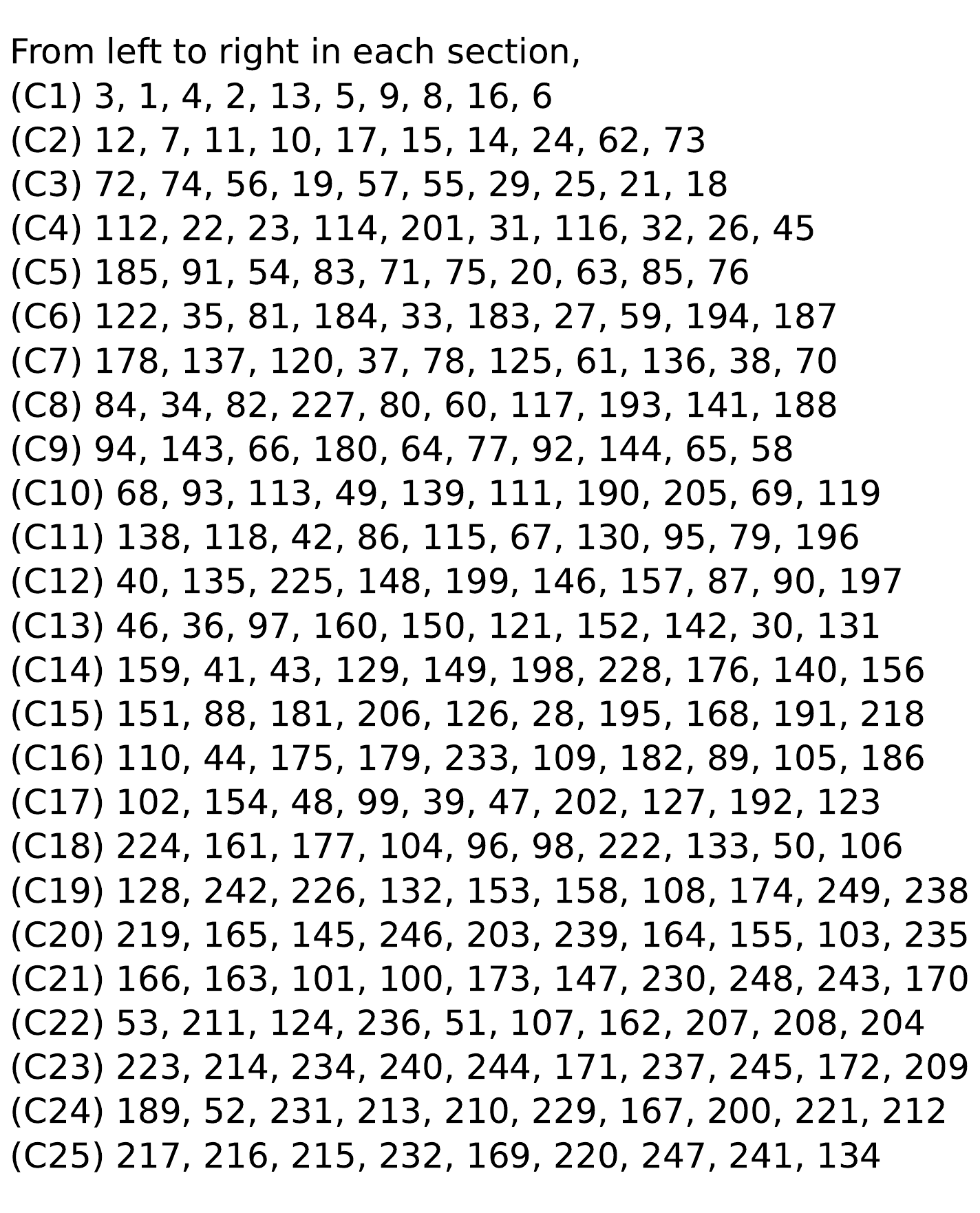}\vspace{3mm}
  \includegraphics[width=0.45\hsize]{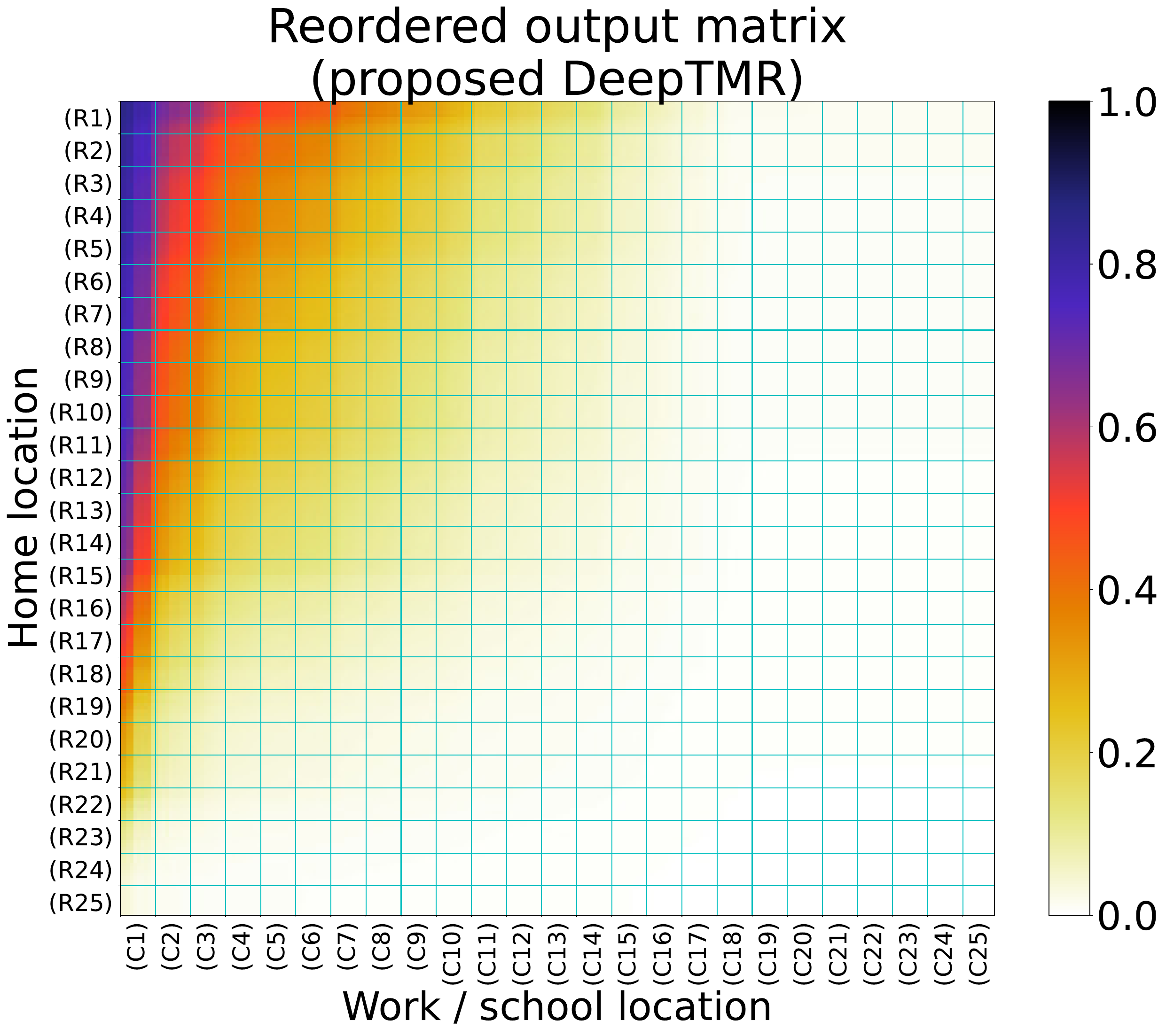}
  \includegraphics[width=0.25\hsize]{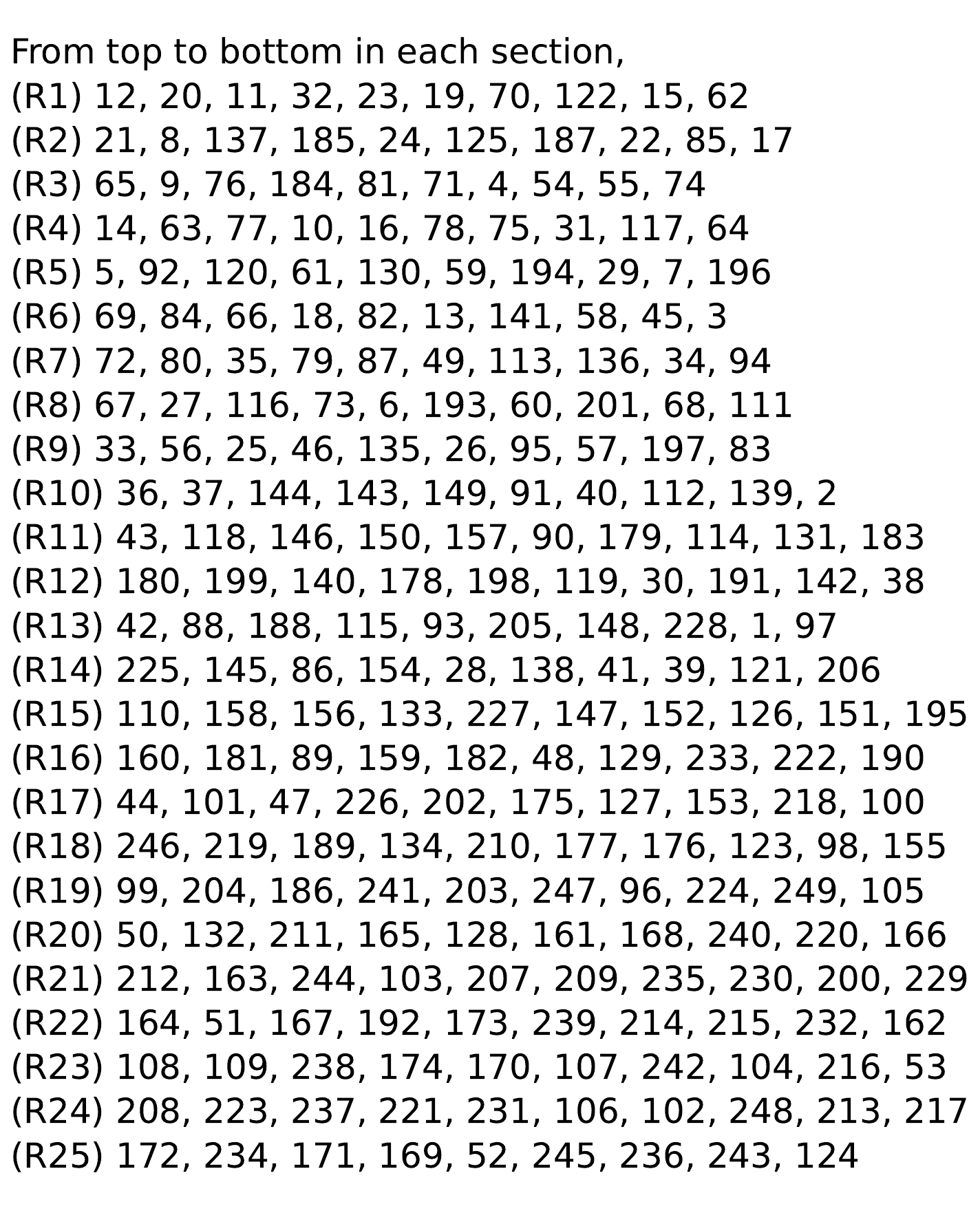}
  \includegraphics[width=0.25\hsize]{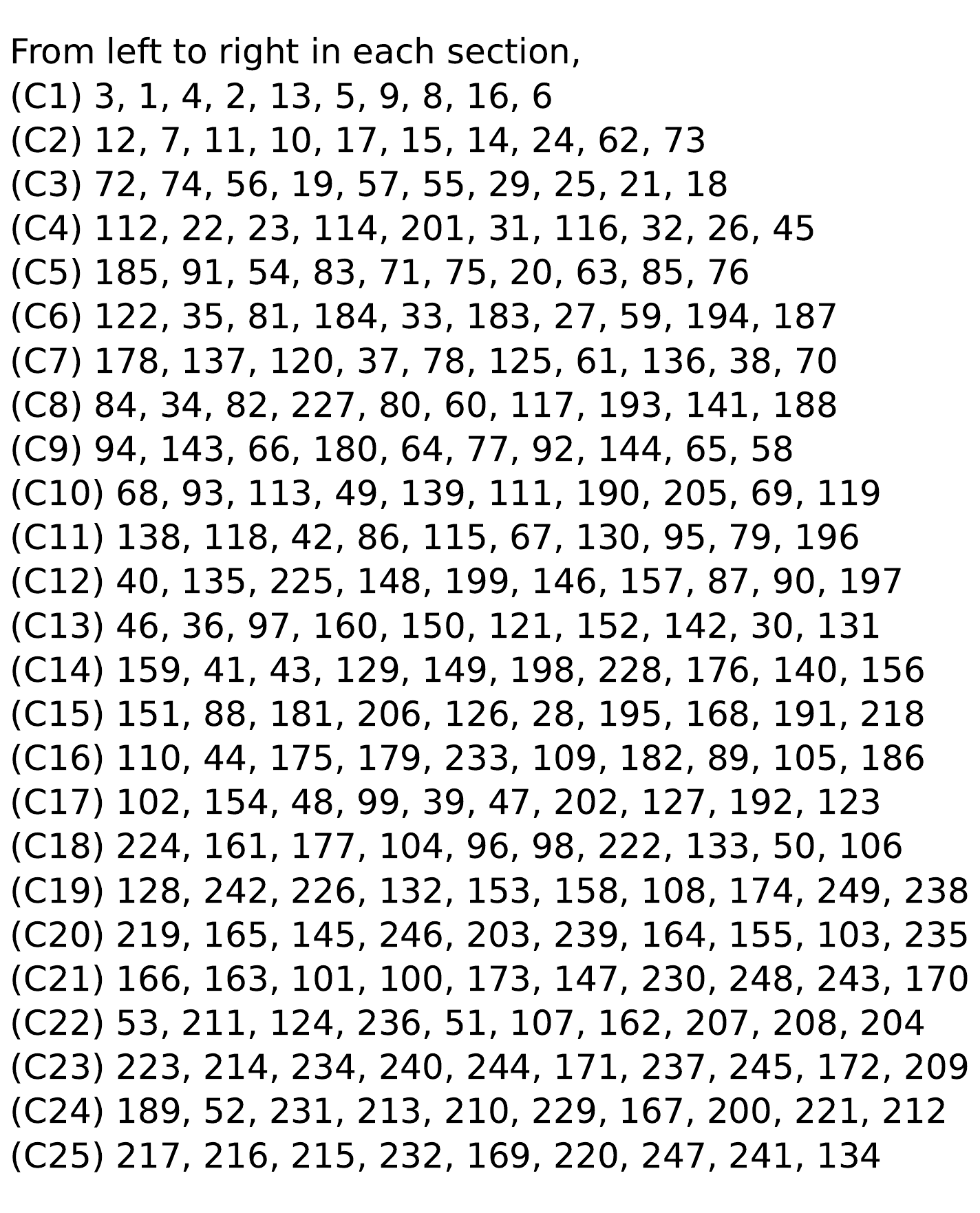}
  \caption{Results of the \textbf{metropolis traffic census dataset} for matrices $A$, $\underline{A}$, and $\underline{\hat{A}}$. For visibility, we plotted the cyan lines to show the sections between the sets of $10$ rows or columns (i.e., $\{ \mathrm{R}1, \dots, \mathrm{R}25 \}$ and $\{ \mathrm{C}1, \dots, \mathrm{C}25 \}$ for rows and columns, respectively). Because the matrix size is $(n, p) = (249, 249)$, $\mathrm{R}25$ and $\mathrm{C}25$ contain nine rows and nine columns, respectively. The correspondence of the indices with the locations is given in Appendix \ref{sec:ap_table}.}\vspace{3mm}
  \label{fig:results_p2_1}
\end{figure}

\begin{figure}[t]
  \centering
  \includegraphics[width=0.45\hsize]{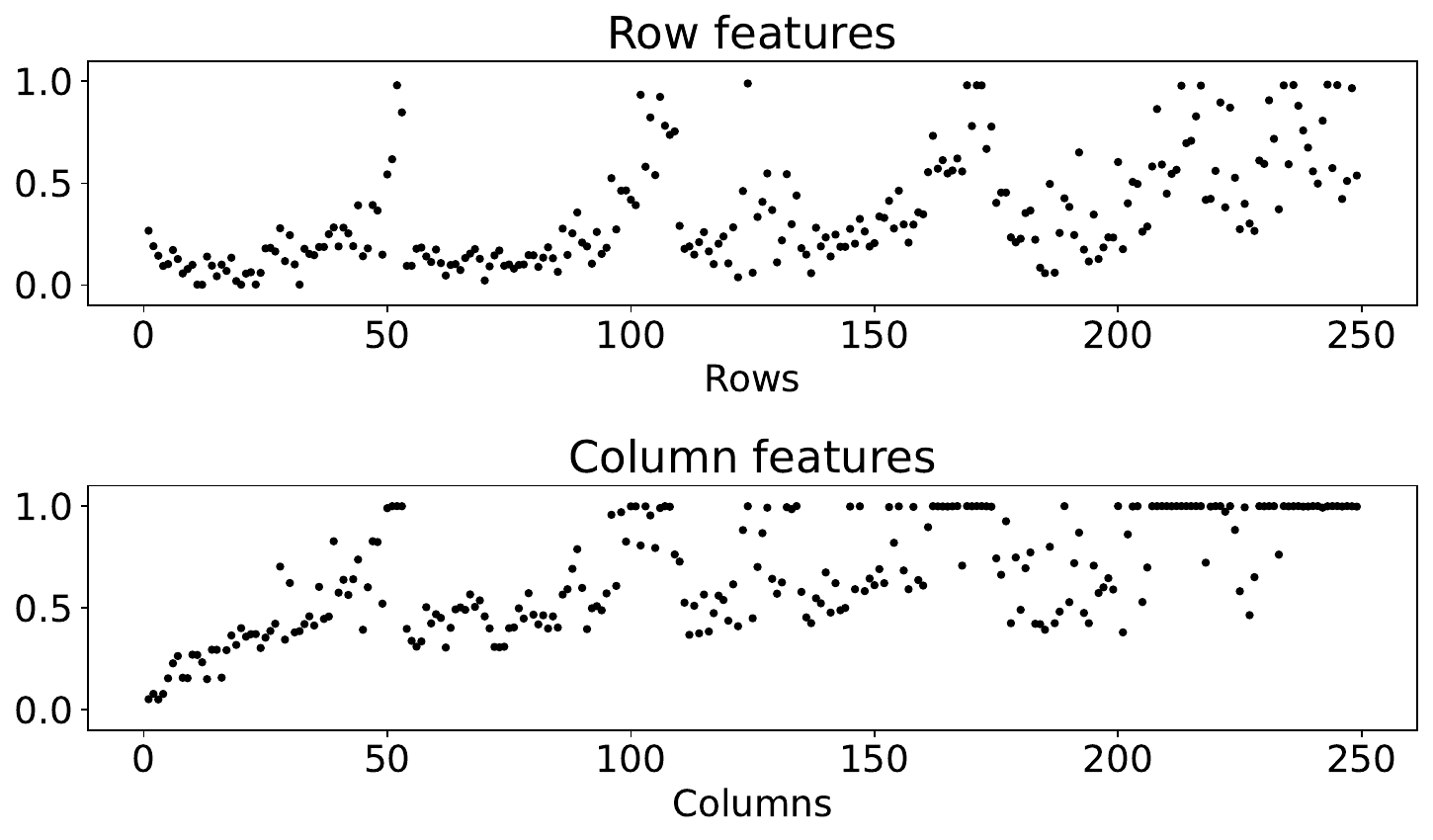}\hspace{10mm}
  \includegraphics[width=0.45\hsize]{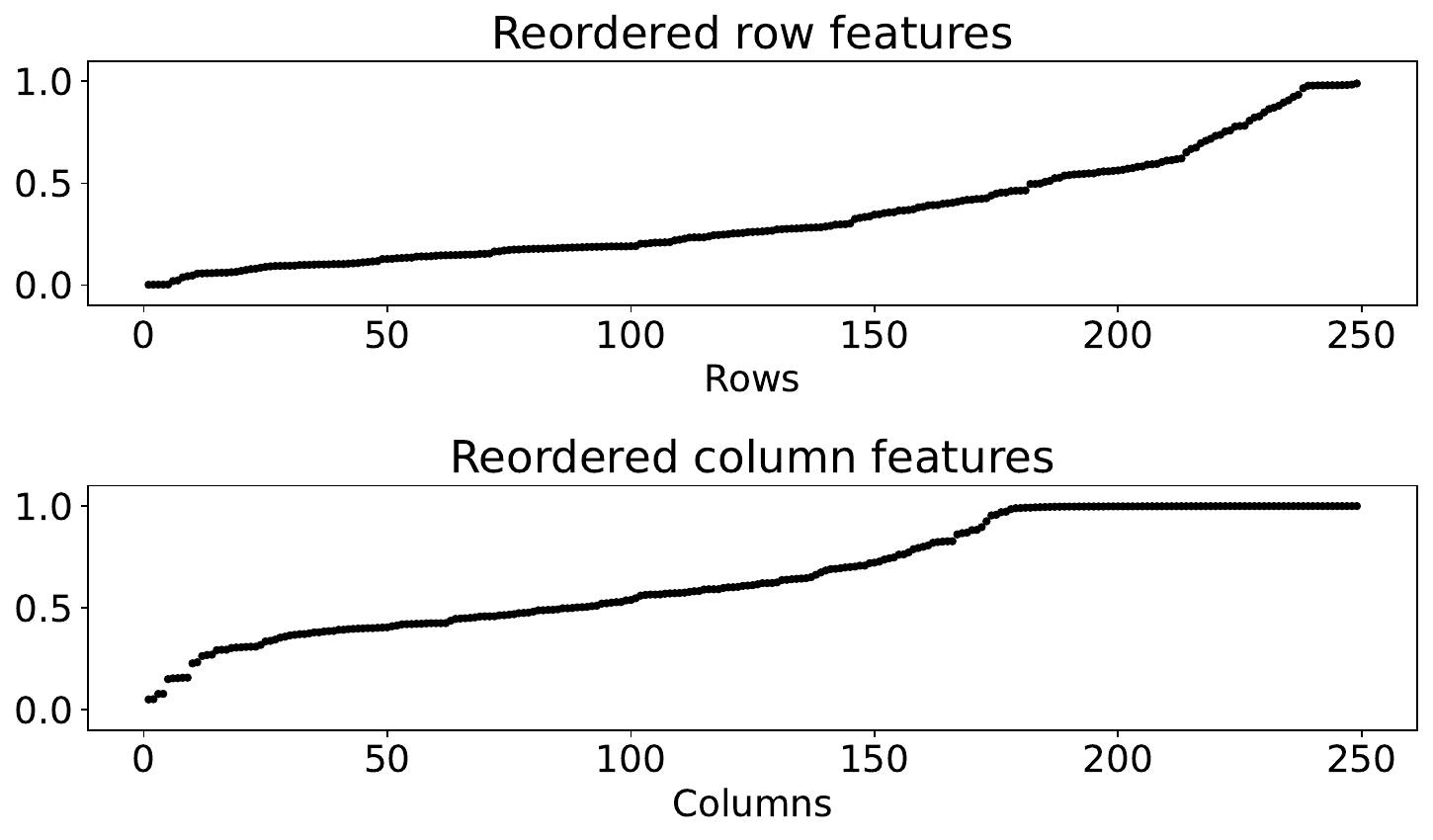}\vspace{-1mm}
  \caption{Results of the \textbf{metropolis traffic census dataset} for vectors $\bm{g}$, $\bm{h}$, $\underline{\bm{g}}$, and $\underline{\bm{h}}$.}\vspace{3mm}
  \label{fig:results_p2_2}
\end{figure}

Figures \ref{fig:results_p2_1} and \ref{fig:results_p2_2} respectively show the results of matrices $A$, $\underline{A}$, $\underline{\hat{A}}$, and vectors $\bm{g}$, $\bm{h}$, $\underline{\bm{g}}$, $\underline{\bm{h}}$ for this dataset. The correspondence of the row and column indices with locations in Figure \ref{fig:results_p2_1} is given in Appendix \ref{sec:ap_table}. The reordered input and output matrices, $\underline{A}$ and $\underline{\hat{A}}$, respectively show that the number of commuting people increases from the lower right to the upper left corner of the matrices. For instance, regardless of the home location, the number of people commuting to the locations in (C21) to (C25) (e.g., Fukaya City in Saitama Prefecture, Tatebayashi City in Gunma Prefecture, and Nogi Town in Tochigi Prefecture) in matrices $\underline{A}$ and $\underline{\hat{A}}$ is relatively small. However, relatively many people commute to locations in (C1) (e.g., Minato-ku, Chiyoda-ku, and Shinjuku-ku in Tokyo), especially from home locations in (R1) -- (R10) (e.g., Setagaya-ku, Nerima-ku, and Ota-ku in Tokyo). 



\section{Discussion}
\label{sec:discussion}

Here, we discuss the results and future research directions. In the experiments in Section \ref{sec:experiments}, we showed that the DeepTMR can successfully reorder both synthetic and practical data matrices and provide their denoised mean matrices as output. Despite its effectiveness, DeepTMR leaves room for further improvement, as described in the subsequent paragraphs. 

One potential merit of the proposed DeepTMR compared to other spectral and dimension-reduction-based methods is that it only requires an $n$-dimensional column data vector and $p$-dimensional row data vector as input, not the entire data matrix. This suggests the possibility that, if a set of rows or columns in a data matrix increases with time, we would not have to train DeepTMR from scratch. 
Instead, we could only fine-tune the previously trained model with newly added data to predict orders. A main problem in realizing this is that the input dimensions of the current DeepTMR should be fixed in advance. However, to apply DeepTMR to such a time-series data matrix, we need to extend the DeepTMR such that it can accept a variable-length input data vector. 
Such an extension is also desirable from the perspective of memory costs. For a large data matrix, a DeepTMR with an $(n + p)$-dimensional input layer requires a large amount of memory to be stored. One possible solution to this problem is to first select randomly $k$ rows and $h$ columns, where $k \ll n$ and $h \ll p$, and use the selected rows and columns as inputs. In this case, we need to develop a model under a different problem setting from ours, where each entry in an input data vector does not necessarily correspond to the same row or column. 

Another limitation of the proposed method is that the trained DeepTMR model is affected by random initialization, mini-batch selection for each iteration, and hyperparameter settings (e.g., number of units in each layer). In the experiment in Section \ref{sec:comparison}, to partially alleviate this problem, we trained the neural network multiple times and chose the result with the minimum training error. However, this na\"{i}ve approach increased the overall computation time. As such, it would be desirable to construct a more sophisticated model that is more robust to the effects of these settings. In particular, it is important to determine the optimal architecture of a neural network for a given data matrix. The experimental results in Section \ref{sec:experiments} show that the DeepTMR could successfully extract the denoised mean matrices of the input matrices with sizes ranging from $100 \times 100$ to $249 \times 249$ by using row/column encoder networks with $10$ units in the middle layer. Based on these results, we expect that we do not always need to set the size of the DeepTMR network as large as the input matrix to extract the structural patterns from a data matrix. 

Finally, it would be interesting to utilize additional input information for the rows and columns besides the entry values of an observed matrix. For instance, in the case of the metropolis traffic census dataset \cite{estat} in Section \ref{sec:traffic_data}, each row or column corresponds to a specific location in Japan. If we can extend the DeepTMR to reorder a data matrix using such additional row/column information (e.g., geographical location), it would be possible to obtain a different structural pattern for the data matrix compared to those provided by the current model. 


\section{Conclusions}
\label{sec:conclusion}

In this paper, we proposed a new matrix reordering method, called DeepTMR, based on a neural network model. By using an autoencoder-like architecture, the proposed DeepTMR can automatically encode the row and column of an input matrix into one-dimensional nonlinear features, which can be subsequently used to determine the row and column orders. Moreover, a trained DeepTMR model provides a denoised mean matrix as output, which illustrates the global structure of the reordered input matrix. Through experiments, we showed that the proposed DeepTMR can successfully reorder the rows and columns of both synthetic and practical datasets and achieve higher accuracy in matrix reordering than the existing spectral and dimension-reduction-based matrix reordering methods based on SVD and MDS. 


\section*{Acknowledgments}

TS was partially supported by JSPS KAKENHI (18K19793, 18H03201, and 20H00576), Japan Digital Design, Fujitsu Laboratories Ltd., and JST CREST. We would like to thank Editage (\url{www.editage.com}) for English language editing.


\clearpage
\begin{appendices}
\section{Correspondence of the attribute indices with meanings in the divorce predictors dataset}
\label{sec:ap_table_divorce}

The meaning of each attribute index of the divorce predictors dataset \cite{Yontem2017, Yontem2019} is as follows: 

\begin{enumerate}
\item If one of us apologizes when our discussion deteriorates, the discussion ends.
\item I know we can ignore our differences, even if things get hard sometimes.
\item When we need it, we can take our discussions with my spouse from the beginning and correct it.
\item When I discuss with my spouse, to contact him will eventually work.
\item The time I spent with my wife is special for us.
\item We don't have time at home as partners.
\item We are like two strangers who share the same environment at home rather than family.
\item I enjoy our holidays with my wife.
\item I enjoy traveling with my wife.
\item Most of our goals are common to my spouse.
\item I think that one day in the future, when I look back, I see that my spouse and I have been in harmony with each other.
\item My spouse and I have similar values in terms of personal freedom.
\item My spouse and I have similar sense of entertainment.
\item Most of our goals for people (children, friends, etc.) are the same.
\item Our dreams with my spouse are similar and harmonious.
\item We're compatible with my spouse about what love should be.
\item We share the same views about being happy in our life with my spouse.
\item My spouse and I have similar ideas about how marriage should be.
\item My spouse and I have similar ideas about how roles should be in marriage.
\item My spouse and I have similar values in trust.
\item I know exactly what my wife likes.
\item I know how my spouse wants to be taken care of when she/he sick.
\item I know my spouse's favorite food.
\item I can tell you what kind of stress my spouse is facing in her/his life.
\item I have knowledge of my spouse's inner world.
\item I know my spouse's basic anxieties.
\item I know what my spouse's current sources of stress are.
\item I know my spouse's hopes and wishes.
\item I know my spouse very well.
\item I know my spouse's friends and their social relationships.
\item I feel aggressive when I argue with my spouse.
\item When discussing with my spouse, I usually use expressions such as ``you always'' or ``you never.''
\item I can use negative statements about my spouse's personality during our discussions.
\item I can use offensive expressions during our discussions.
\item I can insult my spouse during our discussions.
\item I can be humiliating when we discussions.
\item My discussion with my spouse is not calm.
\item I hate my spouse's way of open a subject.
\item Our discussions often occur suddenly.
\item We're just starting a discussion before I know what's going on.
\item When I talk to my spouse about something, my calm suddenly breaks.
\item When I argue with my spouse, I only go out and I don't say a word.
\item I mostly stay silent to calm the environment a little bit.
\item Sometimes I think it's good for me to leave home for a while.
\item I'd rather stay silent than discuss with my spouse.
\item Even if I'm right in the discussion, I stay silent to hurt my spouse.
\item When I discuss with my spouse, I stay silent because I am afraid of not being able to control my anger.
\item I feel right in our discussions.
\item I have nothing to do with what I've been accused of.
\item I'm not actually the one who's guilty about what I'm accused of.
\item I'm not the one who's wrong about problems at home.
\item I wouldn't hesitate to tell my spouse about her/his inadequacy.
\item When I discuss, I remind my spouse of her/his inadequacy.
\item I'm not afraid to tell my spouse about her/his incompetence.
\end{enumerate}

\section{Correspondence of the indices with locations in the metropolis traffic census dataset}
\label{sec:ap_table}

The meaning of each row or column index of the metropolis traffic census dataset \cite{estat} is as follows: 

\begin{itemize}
\item \textbf{[Tokyo]} 1: Chiyoda-ku, 2: Chuo-ku, 3: Minato-ku, 4: Shinjuku-ku, 5: Bunkyo-ku, 6: Taito-ku, 7: Sumida-ku, 8: Koto-ku, 9: Shinagawa-ku, 10: Meguro-ku, 11: Ota-ku, 12: Setagaya-ku, 13: Shibuya-ku, 14: Nakano-ku, 15: Suginami-ku, 16: Toshima-ku, 17: Kita-ku, 18: Arakawa-ku, 19: Itabashi-ku, 20: Nerima-ku, 21: Adachi-ku, 22: Katsushika-ku, 23: Edogawa-ku, 24: Hachioji City, 25: Tachikawa City, 26: Musashino City, 27: Mitaka City, 28: Ome City, 29: Fuchu City, 30: Akishima City, 31: Chofu City, 32: Machida City, 33: Koganei City, 34: Kodaira City, 35: Hino City, 36: Higashimurayama City, 37: Kokubunji City, 38: Kunitachi City, 39: Fussa City, 40: Komae City, 41: Higashiyamato City, 42: Kiyose City, 43: Higashikurume City, 44: Musashimurayama City, 45: Tama City, 46: Inagi City, 47: Hamura City, 48: Akiruno City, 49: Nishitokyo City, 50: Mizuho Town, 51: Hinode Town, 52: Hinohara Village, 53: Okutama Town
\item \textbf{[Yokohama City, Kanagawa Prefecture]} 54: Tsurumi-ku, 55: Kanagawa-ku, 56: Nishi-ku, 57: Naka-ku, 58: Minami-ku, 59: Hodogaya-ku, 60: Isogo-ku, 61: Kanazawa-ku, 62: Kohoku-ku, 63: Totsuka-ku, 64: Konan-ku, 65: Asahi-ku, 66: Midori-ku, 67: Seya-ku, 68: Sakae-ku, 69: Izumi-ku, 70: Aoba-ku, 71: Tsuzuki-ku
\item \textbf{[Kawasaki City, Kanagawa Prefecture]} 72: Kawasaki-ku, 73: Saiwai-ku, 74: Nakahara-ku, 75: Takatsu-ku, 76: Tama-ku, 77: Miyamae-ku, 78: Asao-ku
\item \textbf{[Sagamihara City, Kanagawa Prefecture]} 79: Midori-ku, 80: Chuo-ku, 81: Minami-ku
\item \textbf{[Kanagawa Prefecture]} 82: Yokosuka City, 83: Hiratsuka City, 84: Kamakura City, 85: Fujisawa City, 86: Odawara City, 87: Chigasaki City, 88: Zushi City, 89: Miura City, 90: Hadano City, 91: Atsugi City, 92: Yamato City, 93: Isehara City, 94: Ebina City, 95: Zama City, 96: Minamiashigara City, 97: Ayase City, 98: Hayama Town, 99: Samukawa Town, 100: Oiso Town, 101: Ninomiya Town, 102: Nakai Town, 103: Oimachi, 104: Matsuda Town, 105: Kaisei Town, 106: Hakone Town, 107: Manazuru Town, 108: Yugawara Town, 109: Aikawa Town
\item \textbf{[Saitama City, Saitama Prefecture]} 110: Nishi-ku, 111: Kita-ku, 112: Omiya-ku, 113: Minuma-ku, 114: Chuo-ku, 115: Sakura-ku, 116: Urawa-ku, 117: Minami-ku, 118: Midori-ku, 119: Iwatsuki-ku
\item \textbf{[Saitama Prefecture]} 120: Kawagoe City, 121: Kumagaya City, 122: Kawaguchi City, 123: Gyoda City, 124: Chichibu City, 125: Tokorozawa City, 126: Hanno City, 127: Kazo City, 128: Honjo City, 129: Higashimatsuyama City, 130: Kasukabe City, 131: Sayama City, 132: Hanyu City, 133: Konosu City, 134: Fukaya City, 135: Ageo City, 136: Soka City, 137: Koshigaya City, 138: Warabi City, 139: Toda City, 140: Iruma City, 141: Asaka City, 142: Shiki City, 143: Wako City, 144: Niiza City, 145: Okegawa City, 146: Kuki City, 147: Kitamoto City, 148: Yashio City, 149: Fujimi City, 150: Misato City, 151: Hasuda City, 152: Sakado City, 153: Satte City, 154: Tsurugashima City, 155: Hidaka City, 156: Yoshikawa City, 157: Fujimino City, 158: Shiraoka City, 159: Ina Town, 160: Miyoshi Town, 161: Moroyama Town, 162: Ogose Town, 163: Namegawa Town, 164: Ranzan Town, 165: Ogawa Town, 166: Kawajima Town, 167: Yoshimi Town, 168: Hatoyama Town, 169: Tokigawa Town, 170: Yokoze Town, 171: Higashi Chichibu Village, 172: Misato Town, 173: Kamisato Town, 174: Yorii Town, 175: Miyashiro Town, 176: Sugito Town, 177: Matsubushi Town
\item \textbf{[Chiba City, Chiba Prefecture]} 178: Chuo-ku, 179: Hanamigawa-ku, 180: Inage-ku, 181: Wakaba-ku, 182: Midori-ku, 183: Mihama-ku
\item \textbf{[Chiba Prefecture]} 184: Ichikawa City, 185: Funabashi City, 186: Kisarazu City, 187: Matsudo City, 188: Noda City, 189: Mobara City, 190: Narita City, 191: Sakura City, 192: Togane City, 193: Narashino City, 194: Kashiwa City, 195: Ichihara City, 196: Nagareyama City, 197: Yachiyo City, 198: Abiko City, 199: Kamagaya City, 200: Kimitsu City, 201: Urayasu City, 202: Yotsukaido City, 203: Sodegaura City, 204: Yachimata City, 205: Inzai City, 206: Shiroi City, 207: Tomisato City, 208: Katori City, 209: Sanmu City, 210: Oamishirasato City, 211: Shisui Town, 212: Sakae Town, 213: Kozaki Town, 214: Ichinomiya Town, 215: Chosei Village, 216: Nagara Town, 217: Otaki Town
\item \textbf{[Ibaraki Prefecture]} 218: Tsuchiura City, 219: Koga City, 220: Ishioka City, 221: Yuki City, 222: Ryugasaki City, 223: Shimotsuma City, 224: Joso City, 225: Toride City, 226: Ushiku City, 227: Tsukuba City, 228: Moriya City, 229: Chikusei City, 230: Bando City, 231: Inashiki City, 232: Kasumigaura City, 233: Tsukubamirai City, 234: Miho Village, 235: Ami Town, 236: Kawachi Town, 237: Yachiyo Town, 238: Goka Town, 239: Sakai Town, 240: Tone Town
\item \textbf{[Gunma Prefecture]} 241: Tatebayashi City, 242: Itakura Town, 243: Meiwa Town
\item \textbf{[Tochigi Prefecture]} 244: Tochigi City, 245: Sano City, 246: Oyama City, 247: Nogi Town
\item \textbf{[Yamanashi Prefecture]} 248: Otsuki City, 249: Uenohara City
\end{itemize}

\end{appendices}


\clearpage
\bibliographystyle{abbrv}
\bibliography{paper}

\end{document}